\documentclass[11pt,letterpaper]{article}
\usepackage{epsfig,rotating,setspace,latexsym,amsbsy,amsmath,epsf,amssymb,bm}
\usepackage{cite,graphicx,authblk,color,subfigure}
\usepackage[title]{appendix}
\usepackage{multirow}
\usepackage{ctable}
\usepackage{nicefrac}

\newlength{\Oldarrayrulewidth}

\usepackage{amsmath}
\usepackage{amsthm}
\usepackage{bm}
\usepackage{enumitem}
\usepackage{bbm}

\usepackage{algorithmic}
\usepackage{algorithm}

\usepackage{hyperref}

\newlength\myindent
\newcommand\bindent{%
  \begingroup
  \setlength{\itemindent}{\myindent}
  \addtolength{\algorithmicindent}{\myindent}
}
\newcommand\eindent{\endgroup}

\newtheorem{definition}{Definition}
\newtheorem{theorem}{Theorem}
\newtheorem*{theorem*}{Theorem}
\newtheorem{lemma}{Lemma}

\newtheorem{proposition}{Proposition}

\newtheorem{duplicateprop}{Proposition}
\newtheorem{duplicatelemma}{Lemma}

\title{Fine-Grained Uncertainty Quantification via Collisions}

\author{Jesse Friedbaum, Sudarshan Adiga, Ravi Tandon
\thanks{This work was supported by NSF grants CAREER 1651492, CCF-2100013, CNS-2209951, CNS-1822071, CNS-2317192, and by the U.S. Department of Energy, Office of Science, Office of Advanced Scientific Computing under Award Number DE-SC-ERKJ422, and NIH Award R01-CA261457-01A1. This work was presented in part at the 59th Asilomar Conference on Signals, Systems, and Computers. IEEE, 2025. \cite{friedbaum_asilomar}. The associate editor coordinating the review of this
article and approving it for publication was Yuxin Chen. (Corresponding
author: Ravi Tandon.)}
\thanks{Jesse Friedbaum was with the Department of Electrical and Computer
Engineering, The University of Arizona, Tucson, AZ 85721 USA. He is
now with Pacific Northwest National Laboratory, Richland, WA 99354 USA (e-mail:
jesse.friedbaum@pnnl.gov. 
Sudarshan Adiga was with the Department of Electrical and Computer
Engineering, The University of Arizona, Tucson, AZ 85721 USA. He is
now with Marvell Technology Inc., Santa Clara, CA 95054 USA (e-mail:
sadiga@marvell.com)
Ravi Tandon is with the Department of Electrical and
Computer Engineering, The University of Arizona, Tucson, AZ 85721 USA
(e-mail: tandonr@arizona.edu).
}
}

\begin{document}

\maketitle
\newcommand\blfootnote[1]{%
  \begingroup
  \renewcommand\thefootnote{}\footnote{#1}%
  \addtocounter{footnote}{-1}%
  \endgroup
}


\maketitle

\begin{abstract}
We propose a new and intuitive metric for aleatoric uncertainty quantification (UQ), the prevalence of class collisions defined as the same input being observed in different classes.  We use the rate of class collisions to define the collision matrix, a novel and uniquely fine-grained measure of uncertainty.  For a classification problem involving $K$ classes, the $K\times K$ collision matrix $S$ measures the inherent difficulty in distinguishing between each pair of classes.  
We discuss several applications of the collision matrix,  establish its fundamental mathematical properties, and show its relationship with existing UQ methods, including the Bayes error rate (BER). We also address the new problem of estimating the collision matrix using one-hot labeled data by proposing a series of innovative techniques to estimate $S$. First, we learn a pair-wise contrastive model which accepts two inputs and determines if they belong to the same class. We then show that this contrastive model (which is PAC learnable) can be used to estimate the row Gramian matrix of  $S$, defined as $G=SS^T$.  Finally, we show that under reasonable assumptions, $G$  can be used to uniquely recover  $S$, a new result on non-negative matrices which could be of independent interest.  With a method to estimate $S$ established, we demonstrate how this estimate of $S$, in conjunction with the contrastive model, can be used to estimate the posterior class probability distribution of any point.  Experimental results are also presented to validate our methods of estimating the collision matrix and class posterior distributions on several datasets.
\end{abstract}

\section{Introduction}\label{sec:intro}

\begin{figure*}[t]
\begin{center}
\includegraphics[width=\textwidth]{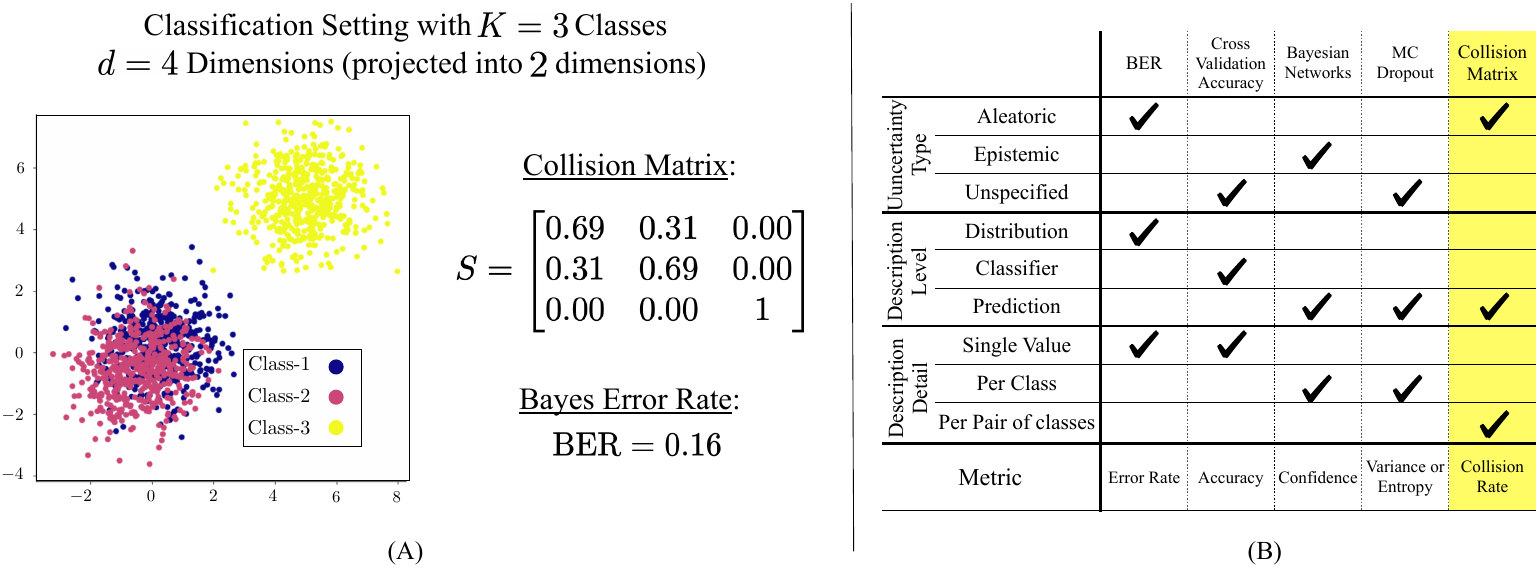}
\caption{\small Part (A) shows the collision matrix ($S$) for a $K=3$ class classification setting where classes $1$ and $2$ are easily confused with each other, but class $3$ is not.  This information is encapsulated in the collision matrix by the values in position $(1,2)$ and $(2,1)$ being larger than the other off-diagonal elements, but this information is not found in the BER.  Part (B) compares the collision matrix against the uncertainty quantification (UQ) measures BER \cite{BER_ensemble}, cross-validation accuracy, Bayesian Networks \cite{BNN_tutorial} and MC Dropout \cite{mc_dropout}.}\label{img:probabilitstic-vs-deterministic}
\end{center}
\vspace{-15pt}
\end{figure*}

As Machine Learning (ML) classifiers are adopted in an ever growing number of fields, they are increasingly being applied in high risk and high uncertainty scenarios such as healthcare \cite{healthcare}, finance \cite{finance}, hiring\cite{hiring} and more.  Many of these settings are inherently probabilistic (contain aleatoric uncertainty), meaning that the observed features do not uniquely determine the class of an input, but rather establish a \textit{posterior probability distribution} over the possible classes.  For example, in healthcare diagnostics, a doctor's observations and the preliminary tests from a patient's first visit may not determine the source of the patient's symptoms with absolute certainty, but these data can determine which ailments are more or less likely and specify the most likely illness. For high-stakes decisions, it is important not only to identify the most likely class but also to understand how likely that classification is to be incorrect and what other classes are the most probable alternatives.  This has led to increased interest in the field of \textit{uncertainty quantification} (UQ), which attempts to describe the uncertainty involved in a prediction task.  Inside the field of UQ, many techniques have been developed which  measure a variety of distinct phenomena using many different metrics (See Figure \ref{img:probabilitstic-vs-deterministic}B).  We propose a new and intuitive metric for uncertainty measurement, \textit{class collisions}, and show how it can be used to describe the general uncertainty in a classification task and the uncertainty associated with individual predictions.  

A class collision is defined as the same input belonging to different classes when observed multiple times.  An example of a class collision in healthcare would be two patients that arrive at a hospital with identical vital signs and complaining of the same symptoms (the input) such as muscle weakness, tingling in the extremities, and blurred vision, but where the symptoms are caused by different diseases (the class) such as Multiple Sclerosis and a Vitamin B12 deficiency.  The prevalence of class collisions between any two classes provides an intuitive characterization of how difficult it is to distinguish between that pair of classes.  We organize this uncertainty data into a statistic called the \textit{collision matrix}, where the $(i,j)^{\text{th}}$ entry of the collision matrix contains the rate of class collisions between classes $i$ and $j$.   In Figure \ref{img:probabilitstic-vs-deterministic}A, we provide an example of a three class setting where classes $1$ and $2$ are easily confused with each other, but class $3$ is not easily confused with any other class.  The collision matrix identifies which pair of classes has uncertainty between them, whereas less detailed UQ measures such as the BER only indicate that there is uncertainty somewhere in the system.   In Section \ref{sec:posterior_estimation}, we show how the collision matrix can also be used to construct posterior class probability distributions of an individual input, i.e., define the probability that an input belongs to each possible class.  This allows us to describe the uncertainty associated with a particular input, whereas the collision matrix itself describes the uncertainty in the whole distribution.

\noindent{\textbf{{UQ Categorization}:}  We now present a categorization of UQ techniques to demonstrate how the collision matrix compares to existing UQ measures  (see Figure \ref{img:probabilitstic-vs-deterministic}A).
\begin{enumerate}[label=\alph*)]
    \item {\textit{Uncertainty Type}}: Uncertainties involved with classification can be broadly divided into two types: \textit{epistemic} and \textit{aleatoric} \cite{AvsE1}.  Epistemic means relating to knowledge and describes the uncertainty resulting from a lack of knowledge, such as insufficient training data or a lack of knowledge about the best model parameters.  For example, Bayesian Networks \cite{Bayesian_Networks} seek to model uncertainty in the choice of model parameters which is epistemic.  Aleatory refers to chance and describes the uncertainty that cannot be reduced with more knowledge.  For example, the Bayes Error Rate or BER \cite{Bayes_optimal} is a measure of aleatoric uncertainty because it refers to the unavoidable error even with access to the best possible classifier.  The collision matrix also measures the aleatoric uncertainty inherent to a data distribution, making it a property of the distribution itself independent of any classifier and would be present even if a full description of the data distribution was available.
    \item {\textit{Description Metric}}: In addition to describing different types of uncertainty, different UQ techniques also use a variety of metrics to measure uncertainty.  Some of these metrics are very intuitive, such as percent confidence or expected error rate, while other measures are more esoteric such as Shannon entropy \cite{UQ_wentropy}, the expected effect on a cost function \cite{label_wise_AandE} or the cardinality of a prediction set \cite{conformal_intro}.  We propose a new and intuitive measure: the expectation that multiple observations of the exact same input values will be found in different classes.  When this expectation is large, it means that uncertainty is very high and no classifier could achieve high accuracy.  On the other hand, when this expectation is zero, it means there is no inherent/aleatoric uncertainty at all---the setting is deterministic.
    \item {\textit{Description Level}}: A UQ measure may describe the uncertainty related to the data distribution itself (such as the BER) or the uncertainty related to a particular classifier (such as the familiar cross-validation accuracy and confusion matrix) or even for an individual prediction, such as in the techniques presented by \cite{bayesian_example}.  Similar to the BER, the collision matrix describes the uncertainty arising from the  distribution itself and not a specific classifier or prediction.  Such measures can be used as an objective standard against which to judge the effectiveness of any classifier.  Although the collision matrix itself describes uncertainty on a distribution level, we present an algorithm in Section \ref{sec:posterior_estimation}, which uses the collision matrix to estimate the posterior class probability distribution, describing the uncertainty involved in a single prediction.
    \item {\textit{Description Detail}}: Some UQ techniques condense all uncertainty into a single value, such as the BER or entropy \cite{UQ_wentropy}.  Other methods, such as \cite{label_wise_AandE}, provide a more detailed description by providing uncertainty values for each class.  The collision matrix provides even more detail by describing uncertainty between each pair of classes.  This level of detail is one of the key advantages of measuring uncertainty with the collision matrix and allows for a fine-grained evaluation of classifier performance.  Because the collision matrix is a property of the data distribution, these uncertainties can be interpreted as measuring the  similarity between each pair of classes.  We also note that combining values in the collision matrix can provide class-wise and single value descriptions of uncertainty as described in the following.
\end{enumerate}}

\begin{figure}[t]
\begin{center}

    \includegraphics[width=0.47\textwidth]{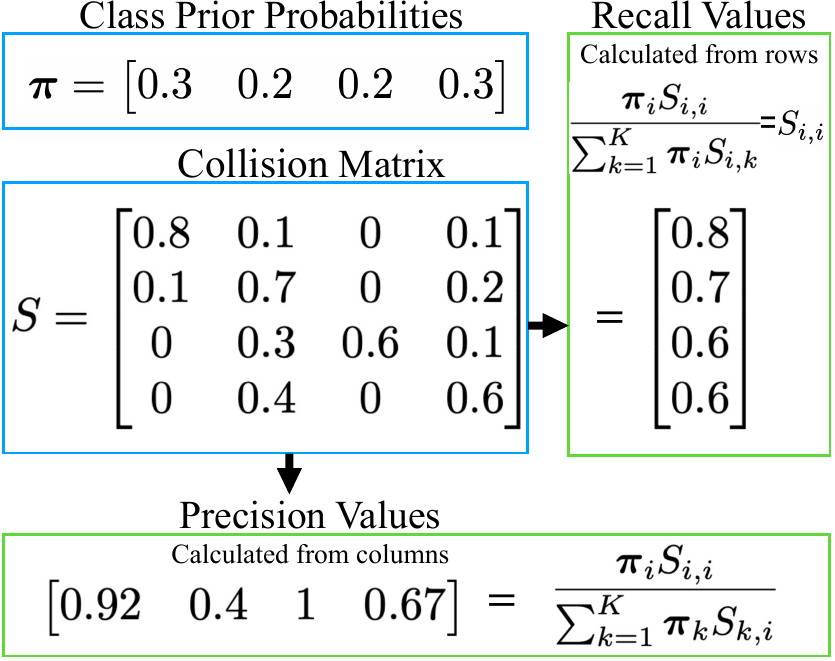}

  \caption{\small Figure demonstrating how the  collision matrix $(S)$ and the class prior probability vector $\boldsymbol{\pi}$ can be used to calculate class precision and recall values. See  Section \ref{sec:BER} for further discussion.}\label{fig:confusion}
\end{center}
\vspace{-15pt}
\end{figure}

\noindent{\textbf{{Deriving further insights from the Collision Matrix}:} In a $K$-class setting, the collision matrix will be a $K\times K$ matrix.  The $(i,j)^{\text{th}}$ entry of this matrix describes the difficulty of distinguishing elements of class $i$ from class $j$, which can be thought of as a measure of uncertainty or an indication of similarity between those classes.  (See an example in Figure \ref{img:probabilitstic-vs-deterministic}A.)   Elements of the collision matrix can also be combined to create less fine-grained measures of uncertainty as illustrated in Figure \ref{fig:confusion}. For instance, the  $i^{\text{th}}$ row of the collision matrix indicates the difficulty of avoiding false negative classifications when the true class is $i$, indicative of the achievable recall on class $i$.  Alternatively, the  $i^{\text{th}}$ column describes how difficult it is to avoid false positive classifications into class $i$, related to the possible precision.  Finally, the diagonal elements of the collision matrix can be used to find a single value measure of the overall uncertainty in $\mathcal{D}$ related to the BER.  We discuss this connection in more detail in Section \ref{sec:BER}.

\vspace{-10pt}

\noindent\textbf{{Applications of the Collision Matrix}:} The collision matrix's ability to quantify uncertainty between pairs of classes makes it a uniquely fine-grained UQ technique and allows for new applications not available to less detailed UQ descriptions.  We introduce a few of these novel applications here. (1) \textit{Class Consolidation:} The collision matrix can be used to predict how the difficulty of a classification scenario will be affected by combining different classes together.  This has the potential to make previously impossible classification problems tractable.  Returning to our healthcare example, it may not be possible to accurately predict the disease from a patient's initial symptoms alone, but the collision matrix could inform the creation of broader disease categories that can be predicted with high accuracy.  (2) \textit{Adaptive Sampling:} Another application of this fine-grained uncertainty measure is to inform online changes to the cost function or sampling strategy during training: if it is observed, during training, that the classifier is very likely to confuse two classes despite there being relatively low uncertainty between these two classes, we could increase the output of the loss function for this error or over-sample from those two classes. In this way the collision matrix does not only indicate how well a classifier has learned a data distribution, but can also indicate how a classifier can be improved.   (3) \textit{Data-Driven Label Smoothing:} Training data is typically \textit{one-hot} labeled, encouraging the classifier to put the entire weight of its prediction on a single class. It has been shown that improved classifiers can be trained by smoothing the labels of training data, that is replacing the one-hot labels with a mixture of the one-hot labels and a uniform distribution \cite{naive_smoothing}.  The collision matrix can be used to smooth labels in a data driven way, replacing the one-hot labels with labels that penalize the classifier less for classifying inputs into classes that are similar to the observed label and penalizing more for assigning an input to a very different class from its label.   (4) \textit{Learning Posterior Distributions:} In  Section \ref{sec:posterior_estimation} we show that the collision matrix can be used to convert similarity scores between an input and each possible class to the posterior class probability distribution of the input.  This application is of particular interest because the posterior class probability distribution fully describes the aleatoric uncertainty relating to a particular input.\newline\newline 


\noindent{\textbf{Summary of Contributions}}:  In Section \ref{sec:defininition}, we rigorously define a novel UQ measure, the collision matrix, as a stochastic matrix comprised of the class collision rates between individual pairs of classes. We  also show how it can be defined using the pdfs describing the data distribution.  Additionally, we show how the collision matrix can be interpreted as the expected confusion matrix of a \textit{Probabilistic Bayes Classifier} which is closely related to the \textit{Bayes Optimal Classifier}, drawing a close link between the collision matrix and the BER.  In Section \ref{sec:G_estimation}, we examine the problem of estimating the collision matrix, developing an estimation method that takes advantage of the unique mathematical properties of the collision matrix.  Specifically, we show that, under contextually justified conditions, the collision matrix is uniquely defined by its row Gramian matrix $G$.  We then show how $G$ can be approximated using a PAC learnable pair-wise contrastive model.  In Section \ref{sec:posterior_estimation}, we show how the collision matrix and the pair-wise contrastive model can be used in conjunction to estimate the posterior class probability distribution of any point.  This allows us to quantify uncertainty for both individual inputs as well as the entire distribution.  Finally, in Section \ref{sec:experiments} we perform numerical evaluations of our methods for estimating the collision matrix and posterior class probability distributions on synthetic and real-world datasets.  We also use calibrated models and Bayesian ensembles to directly estimate class posteriors and the collision matrix.  Our results show that our Gramian based method provides better estimates of the collision matrix, and our method to estimate posteriors using these estimated collision matrices also outperforms baselines.  We show that our method produces sensible results on real-world image data where features of the collision matrix and class posteriors may be intuitively understood. 
Finally, we consider four real-world datasets and show the new/additional insights that one can draw from the collision matrix. 


\section{The Collision Matrix}\label{sec:defininition}

\subsection{Definition of the Collision Matrix}
We consider the setting where training data $\mathcal{T} = \{(\mathbf{x}^{(i)},c^{(i)})\}_{i=1}^n$ is sampled i.i.d. from an unknown distribution $\mathcal{D}$ over ordered pairs $(\mathbf{x},c)$. Here $\mathbf{x}$ represents a $d$ dimensional \textit{feature vector}, $\mathbf{x}\in\mathcal{X} \subset \mathbb{R}^d$.  The label $c$ indicates to which of $K$ classes $\mathbf{x}$ belongs, $c \in \{1,2,...,K\}$.  To further describe $\mathcal{D}$, we use $\boldsymbol{\pi}$ to describe the class prior probabilities, $\boldsymbol{\pi}_k = \mathbb{P}_{\mathcal{D}}(c=k)$.  The conditional distribution of feature vectors in $\mathcal{D}$, given that the class $c=k$ is known, is denoted $\mathcal{D}_k$ with corresponding pdf $f_k(\cdot)$. An object of key importance for understanding uncertainty in a data distribution is the posterior class probability distribution of a feature vector $\mathbf{x}$. 

\begin{definition}[Posterior Class Probability Distribution]\label{def:true_probs}
The posterior class probability distribution of a feature vector $\mathbf{x}$ is defined as 
\begin{align}\label{eqn:true_prob_def}
     \mathbf{y}(\mathbf{x}) := \left(\mathop{\mathbb{P}}_{(\mathbf{x},c)\sim\mathcal{D}}(c=1|\mathbf{x}),\cdot\cdot\cdot , \mathop{\mathbb{P}}_{(\mathbf{x},c)\sim\mathcal{D}}(c=k|\mathbf{x})   \right).
\end{align}
\end{definition}

In general $\mathbf{y}(\mathbf{x})$ can be any element of the $K$-simplex, written as $\mathcal{K}$.  One important special case is when all $\mathbf{y}(\mathbf{x})$ are \textit{one-hot}, meaning $\mathbf{y}_i(\mathbf{x}) =1$  for some $1\leq i\leq K$, and $\mathbf{y}_k(\mathbf{x}) = 0$ for $k\neq i$.  In this case the label is deterministically defined by the feature vector and there is no aleatoric uncertainty in $\mathcal{D}$ (even if individual classifiers may be uncertain in their predictions).  The more the posteriors differ from one-hot, the more uncertainty is found in $\mathcal{D}$. In Section \ref{sec:posterior_estimation} we will present a method to use the collision matrix to estimate these posterior distributions.  Next, we formally define \textit{class collisions} and provide an operational interpretation.

\begin{definition}[Class Collision]\label{def:class_collision}
    A class collision occurs when $(\mathbf{x},c)$ and $(\tilde{\mathbf{x}},\tilde{c})$ are drawn from $\mathcal{D}$ such that $\mathbf{x} = \tilde{\mathbf{x}}$ but $c\neq\tilde{c}$.
\end{definition}

Similar to the class posterior distributions, class collisions can directly indicate whether there is aleatoric uncertainty in a distribution: a distribution $\mathcal{D}$ has aleatoric uncertainty if and only if class collisions are possible.  As the likelihood of class collision increases there is more uncertainty in the distribution, however, the probability of directly observing a class collision, defined by $\mathop{\mathbb{E}}_{(\mathbf{x},c)\sim \mathcal{D}} [ \mathop{\mathbb{E}}_{(\tilde{\mathbf{x}},\tilde{c})\sim \mathcal{D}} [\mathbb{P}(\mathbf{x}=\tilde{\mathbf{x}},c\neq\tilde{c})]]$, is not a good measure of uncertainty.  This is because the odds of ever observing identical inputs in a large feature space $\mathcal{X}$ ($\mathbb{P}(\mathbf{x}=\tilde{\mathbf{x}})$) are almost zero. For example, if a hospital collects variety of information on a new patient (age, height, weight, blood pressure, etc.), it is very unlikely that two patients will ever completely have identical features.  This does not mean, however, that there is little uncertainty in the setting; patient intake data may not be enough to identify a disease with certainty.  For this reason, we use the probability of a collision given that a feature vector has already been observed twice.  We partition these probabilities based on where the input was observed in its first and second appearances to get $K^2$ values.  We organize these values into the \textit{collision matrix}.\newpage

\begin{definition}[Collision Matrix]
    The collision matrix S is a 
    row stochastic matrix defined by
    \begin{equation}\label{eqn:collision_element}
        S_{i,j} := \mathop{\mathbb{E}}_{\mathbf{x}\sim \mathcal{D}_{i}}\mathbb{P}(c=j|\mathbf{x}).
    \end{equation} Each $S_{i,j}$ represents the expectation that a feature vector, which was previously observed in class $i$, will be found in class $j$ if it is observed again, resulting in a class collision.\label{def:confusability_matrix}
\end{definition}

We now present a couple of examples of what the collision matrix $S$ could represent in specific applications.  In our disease diagnosis example, $S_{i,j}$ represents the probability that a patient with the same features as a patient with disease $i$ will actually have disease $j$.  This value indicates how difficult it is to distinguish between the two diseases.  For a second example,  suppose a university gives incoming students a survey about their personality and interests.  The responses to the survey are used as feature vectors with the class corresponding to the major from which the student graduates.  In this example the $S_{i,j}$ represents how often we would expect a student, whose survey answers are identical to a student who graduated in major $i$, to end up graduating from major $j$.  This could be interpreted as describing the inherent difficulty in predicting between those majors or as a measure of the similarity between majors $i$ and $j$ based on the similarity of the students in those majors. 


\subsection{Relation to the Bayes Error Rate (BER)}\label{sec:BER}
We noted in the introduction that the collision matrix is similar to the BER in purpose: both describe the aleatoric uncertainty of the entire distribution $\mathcal{D}$.  We now show how $S$ can further be related to the BER leading to new interpretations of the collision matrix.  The \textit{Bayes Optimal Classifier} is the classifier that achieves the best possible accuracy of any classifier.

\begin{definition}[Bayes Optimal Classifier]
    The Bayes Optimal Classifier $M_{\text{opt}}:\mathcal{X}\rightarrow\{1,2,...,k\}$ always outputs the most likely class
    \begin{align}
        M_{\text{opt}}(\mathbf{x}) := \mathrm{arg}\max (\mathbf{y}(\mathbf{x})).
    \end{align}\label{eqn:bayes_optimal}
\end{definition}

\noindent The BER is simply the error rate of this classifier.  
To link this to the collision matrix we define the closely related \textit{Probabilistic Bayes Classifier} (PBC).
\begin{definition}[Probabilistic Bayes Classifier]\label{def:pbc}
    The Probabilistic Bayes Classifier (PBC) $M_{\text{prob}}:\mathcal{X}\rightarrow\{1,2,...,k\}$ is a non-deterministic classifier that samples its output from the posterior class probabilities
    \begin{equation}
        M_{\text{prob}}(\mathbf{x}) \sim \mathbf{y}(\mathbf{x}).
    \end{equation}\label{eqn:bayes_probable}
\end{definition}

We note that the PBC will predict that an element of class $i$ is in class $j$ at a rate of 
$\mathop{\mathbb{E}}_{\mathbf{x}\sim \mathcal{D}_{i}}\mathbf{y}_j(\mathbf{x}) = \mathop{\mathbb{E}}_{\mathbf{x}\sim \mathcal{D}_{i}}\mathbb{P}(c=j|\mathbf{x}) = S_{i,j}$. 
  This means that the collision matrix can be interpreted as the expected confusion matrix of the PBC.  Accordingly, the off-diagonal elements in column $i$ of $S$ represent the rates of false positives into class $i$, and the off-diagonal elements in row $i$ represent the false negative classification for true class $i$.  
  We can use the diagonal elements of $S$ to produce a single value measure of uncertainty by finding the error rate of the PBC which we denote as the PBER.   Figure \ref{fig:confusion} shows how $S$ can be used to find precision, recall.
\begin{equation}\label{eqn:PBER}
    \text{PBER}(\mathcal{D}) := \sum_{k=1}^K \boldsymbol{\pi}_k(1-S_{k,k})= 1- \sum_{k=1}^{K}\boldsymbol{\pi}_k S_{k,k}.
\end{equation}
For the special case of uniform class priors, i.e., $\boldsymbol{\pi}_k= 1/K$ for all $k$, the above expression simplifies to $\text{PBER}(\mathcal{D})= 1- \text{Tr}(S)/K$, where $\text{Tr}(S)$ denotes the trace of the collision matrix $S$.

\subsection{Additional Properties and Case Study}\label{case_study}

We now examine the mathematical properties of $S$, which are described in Proposition \ref{thm:collision_properties}.  We will then analyze the properties of the collision matrix in the $K=2$ setting to gain additional insights. 
\begin{proposition}[Properties of $S$]
    The entries of the collision matrix $S$ are defined by
    \begin{equation}
        S_{i,j} = \mathop{\mathbb{E}}_{\mathbf{x}\sim \mathcal{D}_{i}}\mathbb{P}(C=j|\mathbf{x}) = \boldsymbol{\pi}_{j}\int_{\mathcal{X}}\frac{f_{{i}}(\mathbf{x})f_{{j}}(\mathbf{x})}{\sum_{k=1}^K \boldsymbol{\pi}_{k} f_{{k}}(\mathbf{x})}d\mathbf{x}.\label{eqn:element_expanded}
    \end{equation}
    $S$ is a row stochastic matrix. In the special case with uniform class priors, i.e., $\boldsymbol{\pi} = \left(\frac{1}{K},\frac{1}{K},...,\frac{1}{K}\right)$, 
    $S$ is a symmetric matrix and, therefore, doubly stochastic.\label{thm:collision_properties}
\end{proposition}
\noindent The proof of Proposition \ref{thm:collision_properties} is found in the Appendix \ref{app:s_properties}.

\noindent\textbf{{Case Study for $K=2$ \& Collision Divergence}:} To better understand the implications of Proposition \ref{thm:collision_properties}, we examine the simplest non-trivial case of binary classification ($K=2$) with equal priors ($\boldsymbol{\pi} = \left(\frac{1}{2},\frac{1}{2}\right)$), which arises in several practical scenarios. In this case, the $2\times 2$ collision matrix $S$ is doubly stochastic and fully described by the off diagonal entry 
\begin{equation}\label{eqn:2by2}
    S_{1,2}=\int_{\mathcal{X}}\frac{f_{\mathcal{D}_1}(\mathbf{x}) f_{\mathcal{D}_2}(\mathbf{x})}{f_{\mathcal{D}_1}(\mathbf{x}) + f_{\mathcal{D}_2}(\mathbf{x})}d\mathbf{x},
\end{equation}
which measure the similarity of $\mathcal{D}_1$ and $\mathcal{D}_2$.  We note that we can lower bound $S_{1,2}\geq0$, because each term in \eqref{eqn:2by2} is non-negative.  Furthermore, we may prove an upper bound for $S_{1,2}$ through the use of the arithmetic mean harmonic mean inequality, which stipulates 
\begin{align}
\frac{2 f_{\mathcal{D}_1}(\mathbf{x}) f_{\mathcal{D}_2}(\mathbf{x})}{f_{\mathcal{D}_1}(\mathbf{x}) + f_{\mathcal{D}_2}(\mathbf{x})} \leq \frac{f_{\mathcal{D}_1}(\mathbf{x}) + f_{\mathcal{D}_2}(\mathbf{x})}{2}.
\label{eq: AM-HM inequality}
\end{align}
Substituting \eqref{eq: AM-HM inequality} in \eqref{eqn:2by2}, we have that
\begin{equation}
S_{1,2}\leq\int_{\mathcal{X}}\frac{f_{\mathcal{D}_1}(\mathbf{x}) + f_{\mathcal{D}_2}(\mathbf{x})}{4}d\mathbf{x}=\frac{1}{4}\left ( \int_{\mathcal{X}}f_{\mathcal{D}_1}(\mathbf{x})d\mathbf{x} + \int_{\mathcal{X}}f_{\mathcal{D}_2}(\mathbf{x})d\mathbf{x} \right ) = \frac{1}{2}.
\end{equation}
\begin{figure}[t]\label{fig:C-Distance}
\begin{center}
    \includegraphics[width=0.6\textwidth]{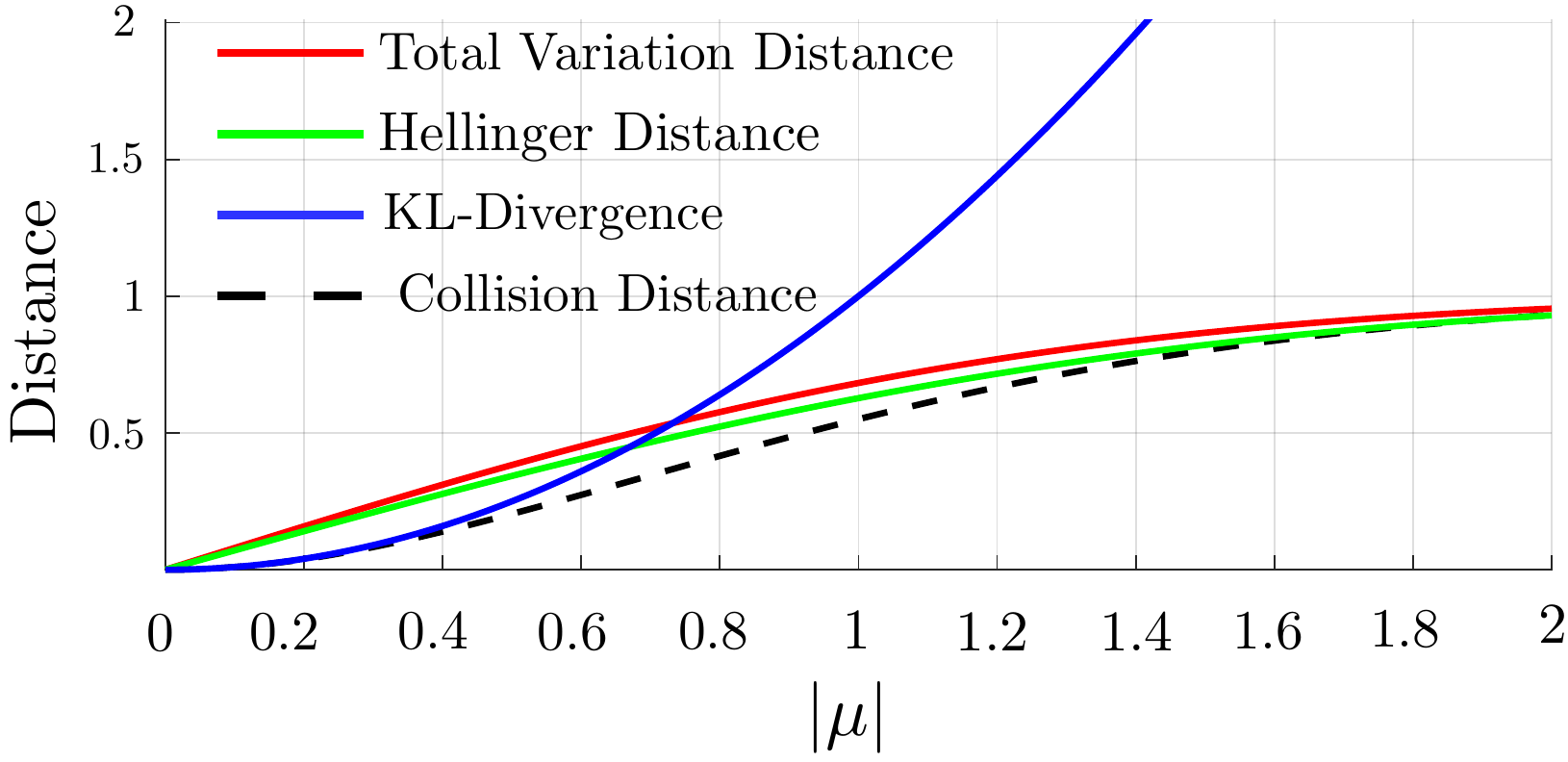}
  \caption{\small A comparison of $\mathcal{D}_{\text{collision}}$ (Collision Divergence) to other common statistical divergences. We compare the divergence between $2$ normal distributions with variance $\sigma^2=1$ means $\mu$ and $-\mu$.}
\end{center}
\end{figure}

We have now shown  that $S_{1,2}\in[0,1/2]$, and we further note that the lower bound of $0$ is achieved when $\mathcal{D}_1$ and $\mathcal{D}_2$ have disjoint support (are totally distinct). On the other hand, the upper bound of $1/2$ is achieved only when $\mathcal{D}_1$ and $\mathcal{D}_2$ are identical.  We can then use $S_{1,2}$ as a measure of how similar or dissimilar two distributions are based on the likelihood of class collisions.  In order to match commonly used measures of statistical divergence such as KL-divergence or Total-Variation distance, however, we must shift and rescale $S_{1,2}$ so that lower values indicate greater similarity.  This results in the following definition.
\begin{definition}[Collision Divergence]
The Collision Divergence between two distributions $\mathcal{D}_1$ and $\mathcal{D}_2$ over the same sample space is defined as 
\begin{equation}
    D_{\text{collision}}(\mathcal{D}_1,\mathcal{D}_2) := 1-2S_{1,2} = 1-\int_{\mathcal{X}}\frac{2f_{\mathcal{D}_1}(\mathbf{x}) f_{\mathcal{D}_2}(\mathbf{x})}{f_{\mathcal{D}_1}(\mathbf{x}) + f_{\mathcal{D}_2}(\mathbf{x})}d\mathbf{x}
\end{equation}    
\end{definition}
 In Figure \ref{fig:C-Distance} we compare $D_{\text{collision}}$ to common statistical divergence measures for univariate Gaussian distributions $\mathcal{N}(\mu,1)$ and $\mathcal{N}(-\mu,1)$ as a function of $\mu$. Remarkably, $D_{\text{collision}}$ exhibits a unique shape with two inflection points that is distinct from Total-Variation, Hellinger and KL-divergences.  

 The Collision Divergence can also be practically interpreted in terms of the Probabilistic Bayes Error Rate or PBER from \eqref{eqn:PBER}.  Indeed, in the binary classification setting with equal priors, $S_{1,2}$ is equal to the PBER.  In this context, the Collision Divergence can be interpreted as how close the PBC comes to achieving its best possible performance on a binary classification problem with uniform prior.  For example, when $D_{\text{collision}}(\mathcal{D}_1,\mathcal{D}_2)=0$ it indicates that the PBC achieves it's worst possible accuracy of $0.5$, and it is as challenging as possible for the PBC to distinguish $\mathcal{D}_1$ and $\mathcal{D}_2$.  Alternatively, when $D_{\text{collision}}(\mathcal{D}_1,\mathcal{D}_2)=1$, the PBC achieves its maximum accuracy of $1$ and it is as easy as possible for the PBC to differentiate $\mathcal{D}_1$ and $\mathcal{D}_2$.

 We can also draw a parallel between the familiar KL-Divergence and the Collision Divergence.  One practical interpretation of KL-divergence $D_{\text{KL}}(\mathcal{D}_1||\mathcal{D}_2)$ is the increase in the expected number of bits required to code a message if we use the optimal encoding for $\mathcal{D}_2$ instead of $\mathcal{D}_1$, and the source of the message is distributed according to $\mathcal{D}_1$.  This can be directly observed by rewriting the formula for KL-divergence as a difference of expectations
 \begin{align}
     D_{\text{KL}}(\mathcal{D}_1||\mathcal{D}_2) =& \int_\mathcal{X}f_{\mathcal{D}_1}(\mathbf{x})\ln\left (\frac{f_{\mathcal{D}_1}(\mathbf{x})}{f_{\mathcal{D}_2}(\mathbf{x})}\right )d\mathbf{x}\notag\\
     =&\left(-\int_\mathcal{X}f_{\mathcal{D}_1}(\mathbf{x})\ln(f_{\mathcal{D}_2}(\mathbf{x}))d\mathbf{x}\right)-\left (-\int_\mathcal{X}f_{\mathcal{D}_1}(\mathbf{x})\ln(f_{\mathcal{D}_1}(\mathbf{x}))d\mathbf{x} \right ),\label{eqn:kl_interp} 
 \end{align}
 where the first term in \eqref{eqn:kl_interp} is the expected code length using the optimal encoding for $\mathcal{D}_2$ and the second term is the expected code length using the optimal encoding for $\mathcal{D}_1$.  Similarly, the Collision Divergence can be interpreted as the decrease in the likelihood of a of point belonging to class $2$ instead of class $1$ when the point is drawn from distribution $\mathcal{D}_1$.  This can also be shown by rewriting the collision distance as a difference of expectation
 \begin{align}
     D_{\text{collision}}(\mathcal{D}_1,\mathcal{D}_2) =& \int_\mathcal{X}f_{\mathcal{D}_1}(\mathbf{x})d\mathbf{x}-\int_{\mathcal{X}}\frac{2f_{\mathcal{D}_1}(\mathbf{x}) f_{\mathcal{D}_2}(\mathbf{x})}{f_{\mathcal{D}_1}(\mathbf{x}) + f_{\mathcal{D}_2}(\mathbf{x})}d\mathbf{x}\notag\\
     =&\int_{\mathcal{X}}\left (\frac{f^2_{\mathcal{D}_1}(\mathbf{x}) + f_{\mathcal{D}_1}(\mathbf{x})f_{\mathcal{D}_2}(\mathbf{x})}{f_{\mathcal{D}_1}(\mathbf{x}) + f_{\mathcal{D}_2}(\mathbf{x})}-\frac{2f_{\mathcal{D}_1}(\mathbf{x}) f_{\mathcal{D}_2}(\mathbf{x})}{f_{\mathcal{D}_1}(\mathbf{x}) + f_{\mathcal{D}_2}(\mathbf{x})}\right )d\mathbf{x}\notag\\
     =&\int_{\mathcal{X}}f_{\mathcal{D}_1}(\mathbf{x})\frac{f_{\mathcal{D}_1}(\mathbf{x})}{f_{\mathcal{D}_1}(\mathbf{x}) + f_{\mathcal{D}_2}(\mathbf{x})}d\mathbf{x}-\int_{\mathcal{X}}f_{\mathcal{D}_1}(\mathbf{x})\frac{f_{\mathcal{D}_2}(\mathbf{x})}{f_{\mathcal{D}_1}(\mathbf{x}) + f_{\mathcal{D}_2}(\mathbf{x})}d\mathbf{x}.\label{eqn:coll_interp}
 \end{align}
 The first term in \eqref{eqn:coll_interp} is the expected likelihood of a input drawn from $\mathcal{D}_1$ belonging to class $1$ and the second term is is the expected likelihood of an input drawn from $\mathcal{D}_1$ belonging to class $2$.  Despite this similarity in interpretation, the Collision Divergence behaves very differently form KL Divergence because a difference in likelihoods is bounded by $1$, whereas, a difference in code lengths can be unbounded (see Figure \ref{fig:C-Distance}).  Now that we have explored the mathematical properties of the collision matrix, we turn our attention to how $S$ may be estimated from the training data.

\section{Estimating the Collision Matrix}
\label{sec:G_estimation}

In this section we will introduce a novel method for estimating the collision matrix by first estimating the  row Gramian of the collision matrix and then recovering $S$ from its row Gramian. 
\begin{itemize}
    \item \textbf{Step 1:} Estimate the row Gramian $G=SS^T$
    \item \textbf{Step 2:} Recover $S$ from $G$
\end{itemize}
This technique is made possible by unique mathematical properties of $S$.  Before we introduce this method, however, we discuss baseline methods that can be used to estimate $S$ without taking advantage of its special mathematical properties.

\noindent\textbf{{Baseline Approaches}:}  One approach for estimating the collision matrix is to replace the expectation in equation \eqref{eqn:collision_element} with an empirical mean.  If we partition the training data by the class of each point, $\mathcal{T}_k = \{\mathbf{x}| (\mathbf{x},c)\in \mathcal{T}, c = k\}$, this approach becomes:
\begin{equation}\label{eqn:naive_approx}
    S_{i,j} \approx \frac{1}{|\mathcal{T}_i|}\sum_{x\in\mathcal{T}_i}\mathbb{P}(c=j|\mathbf{x}) = \frac{1}{|\mathcal{T}_i|}\sum_{x\in\mathcal{T}_i}\mathbf{y}_{j}(\mathbf{x}).
\end{equation}  Solving \eqref{eqn:naive_approx}, however, requires the class posterior probabilities $\{\mathbf{y}(\mathbf{x}^{(i)})\}_{i=1}^n$ (which are not given) and does not utilize the unique mathematical properties of $S$.  Note that classifiers $M:\mathcal{X}\rightarrow\mathcal{K}$ are typically trained to maximize accuracy and $M(\mathbf{x})$ may not be close to $\mathbf{y}(\mathbf{x})$.  It is possible to use existing UQ techniques to improve estimates of the posterior distributions such as \textit{calibrating} the model \cite{calibration_nn} or creating a distribution of classifier outputs through \textit{Monte Carlo (MC) Dropout} \cite{mc_dropout} or a \textit{Bayesian network} \cite{BNN_tutorial}. See more details on how these methods can be used to estimate posteriors in Section \ref{sec:prior_posteriors}.  In Section \ref{sec:experiments}, we use \eqref{eqn:naive_approx} and these methods as baselines for estimating $S$.  

We next describe a novel two step method for estimating $S$ by estimating the row Gramian of the collision matrix and then discovering $S$ from its row Gramian.  Note that this method avoids estimates of $\{\mathbf{y}(\mathbf{x}^{(i)})\}_{i=1}^n$ by taking advantage of the mathematical properties of $S$.  We show that our method outperforms the baselines in Section \ref{sec:experiments}.



\subsection[Recoverability of S from G:]{Recoverability of $S$ from $G$:}\label{sec:recoverablility}  Our two step method involves first estimating the row Gramian matrix of $S$, defined as $G=SS^T$, and then recovering $S$ from $G$.  To justify this we must show that the $G$ uniquely determines $S$.   
In general, for any row Gramian matrix $G$ there are infinitely many matrices $L$ satisfying $LL^T=G$ (see Lemma \ref{lemma:unique_ortho}).  In our specific application to collision matrices, however, there are reasonable restrictions to the type the matrix $S$ satisfying $SS^T=G$ that ensure that the collision matrix $S$ is uniquely defined by $G$.  Specifically, these restrictions are:
\begin{itemize}
    \item \textit{Non-Negativity:}  Any collision matrix $S$ will be non-negative by definition.
    \item \textit{Symmetry:}  Proposition \ref{thm:collision_properties} guarantees $S$ be symmetric when the prior distribution $\boldsymbol{\pi}$ is uniform.
    \item \textit{Strict Diagonal Dominance:} In reasonable classification scenarios, $S$ is diagonally dominant.
\end{itemize}
In the remark below, we explain why the diagonal dominance condition is reasonable in the particular circumstance of a collision matrix, but first we note that the symmetry condition turns the problem of recovering $S$ from $G$ into a matrix square root problem, i.e. {the problem of finding a matrix $S$ such that $SS=G$ when $G$ is given}.  Matrix square roots and their uniqueness, including the case of stochastic matrices (such as the collision matrix), have been studied previously \cite{stochastic_roots}.  To the best of our knowledge, however, the uniqueness of the matrix square root under the specific condition listed above has never been investigated.  In Theorem \ref{thm:unique_root}, we provide a new theoretical result that guarantees strictly diagonally dominant non-negative symmetric matrix square roots are unique.  First, however, we justify the diagonal dominance assumption.

\noindent\textbf{Remark on Diagonal Dominance:} The requirement that a collision matrix $S$ be strictly diagonally dominant,
    \begin{equation}\label{eqn:daigonal_dominance}
        |S_{i,i}| > \sum_{k\neq i}|S_{i,k}|
    \end{equation}
    for all $1\leq i \leq K$, is equivalent to 
    \begin{equation}\label{eqn:dominance_interpretation}
        \mathop{\mathbb{E}}_{\mathbf{x}\sim \mathcal{D}_{i}}\mathbb{P}(c=i|\mathbf{x}) > \mathop{\mathbb{E}}_{\mathbf{x}\sim \mathcal{D}_{i}} \mathbb{P}(c\neq i|\mathbf{x})
    \end{equation}
    for all $1\leq i \leq K$.  The practical interpretation of \eqref{eqn:dominance_interpretation} is that  feature vectors drawn from the distribution of inputs in class $i$ are---in expectation---more likely to be in class $i$ than the other class.  This is not necessarily true for every $\mathcal{D}$, especially when there are large disparities in class priors ($\boldsymbol{\pi}_i<<\boldsymbol{\pi}_j$ for some $i$ and $j$), but we argue that $\mathcal{D}$ does not represent a reasonable classification scenario when \eqref{eqn:dominance_interpretation} is not satisfied.  
    Consider the case where \eqref{eqn:dominance_interpretation} is not satisfied:  there would exist a class $i$ such that, even if we knew precise distribution of inputs in class $i$ ($\mathcal{D}_i$), sampling from this distribution would yield points outside class $i$ for the majority of samples.  With that level of uncertainty, any attempts to classify points into class $i$ would be futile.  Accordingly, in any tractable classification setting, we would expect diagonal dominance to be satisfied.  
    
    \noindent \textbf{Remark on Symmetry:} In the case where the prior distribution $\boldsymbol{\pi}$ is not uniform,  it is possible that other matrices $L\neq S$ exist such that $LL^T = G$.  This means that, in Step $2$ of our Gramian based method (recover $S$ from $G$), we may inadvertently return one such $L$ instead of the true collision matrix.  This does not imply that the Gramian based method will not return the correct collision matrix and in Section $3.3$ we will present a heuristic to help return the correct collision matrix.  We now present our theorem proving that $G$ uniquely determines $S$ under the restrictions of symmetry, strict diagonal dominance, and non-negativity.

 \begin{theorem}[Uniqueness of Row Gramian Factorization]
     Let $G=SS^{T}$ be the row Gramian matrix of a strictly diagonally dominant non-negative symmetric matrix $S$.  Then $S$ is the only diagonally dominant non-negative symmetric matrix satisfying $SS^T=G$. \label{thm:unique_root}
 \end{theorem}


In order to prove Theorem \ref{thm:unique_root}, we will make use of the \textit{Gerschgorin Circle Theorem} \cite{gerschgorin} and a lemma about what type of matrices may share a row Gramian in the general case.  We present both of these results below.

\begin{theorem*}[Gerschgorin Circle Theorem]
     Let $A$ be a $K\times K$ (possibly complex) matrix.  For every $1\leq i \leq K$, define 
     \begin{equation}R(A)_i:=\sum_{k \neq i}|A_{ik}|,\end{equation}
     the absolute sum of the off-diagonal entries in the $i^{\text{th}}$ row.  Further, define the Gerschgorin disks as
     \begin{equation}
         D_i(A) := \{x\in \mathbb{C}||R(A)_i\geq |x-A_{ii}|\}.
     \end{equation}
     That is, each $D_i(A)$ is the closed disk centered at $A_{ii}$ with radius $R_i$.  Then every eigenvalue of $A$ lies within at least one of the Gerschgorin discs $D_i(A)$.
     
 \end{theorem*}

 \begin{lemma}\label{lemma:unique_ortho}
    For any two real square matrices $A$ and $B$, $AA^T=BB^T$ if and only if there exists an orthogonal matrix $Q$, such that $A=BQ$.
\end{lemma}


The Gerschgorin Circle Theorem is a well known result, and a proof may be found in \cite{LinAlg_book} where it is Theorem 6.1.1.  Lemma \ref{lemma:unique_ortho} is a fairly standard linear algebra result, however, we also provide a concise proof of this lemma in Appendix \ref{app:general_gramian}.  With these results in hand, we may now complete our proof.\newline

\noindent\textbf{Proof of Theorem \ref{thm:unique_root}:} Suppose that $L$ is a diagonally dominant non-negative symmetric matrix satisfying $LL^T=G$.  We will prove that $L=S$.  We will first use the Gerschgorin Circle Theorem to show that $S$ and $L$ are both positive definite.  We will next use Lemma \ref{lemma:unique_ortho} to show what $S$ and $L$ commute and are, therefore, simultaneous diagonalizable.  Finally, we use the facts that $S$ and $L$ are both positive definite, simultaneously diagonalizable, and  $SS^T=LL^T$ to show that $S=L$.\newline

\noindent\textit{Positive Definiteness:} We begin by showing that both $S$ and $L$ are positive definite.  Note that, because $S$ and $L$ are non-negative and strictly diagonally dominant, we have
\begin{align}
    S_{ii} &> \sum_{k\neq i}S_{ik} = R(S)_i\geq 0\label{eqn:S_right_half}\\
    L_{ii} &> \sum_{k\neq i}L_{ik} = R(L)_i\geq 0\label{eqn:L_right_half}
\end{align}
for all $1\leq i\leq K$.  Recall that the Gerschgorin disks $D_i(S)$ and $D_i(L)$ are centered at $S_{ii}$ and $L_{ii}$ and have radii $R(S)_i$ and $R(L)_i$ respectively.  Inequalities \eqref{eqn:S_right_half} and \eqref{eqn:L_right_half} then imply that all the Gerschgorin disks lie entirely in the right half-plane.  Because all eigenvalues lie within at least one Gerschgorin disk, this implies that all the eigenvalues have positive real part.  Furthermore, because $S$ and $L$ are symmetric, they have real eigenvalues and every eigenvalue of $S$ and $L$ is a positive real number.  This implies both $S$ and $L$ are positive definite matrices.\newline

\noindent\textit{$S$ and $L$ Commute:}   First note that $LL = LL^T = SS^T=SS$.  By Lemma \ref{lemma:unique_ortho} we have that $L = SQ$ for some orthogonal $Q$.  Because $S$ is positive definite, it is invertible and we may write $Q = S^{-1}L$.  Then $Q$ is the product of symmetric matrices and is symmetric itself.  Furthermore, because $Q$ is symmetric and orthogonal, it is involutory or its own inverse: $QQ = I$.  We may use this to show
 \begin{align}
     L &= SQ\label{eqn:basic_connection}\\
     LQ &= SQQ\\
     LQ &= S.\label{eqn:mod_connection}
 \end{align}
 We may now show that $S$ and $L$ commute, starting by multiplying $S$ to both sides of \eqref{eqn:basic_connection}:
 \begin{align}
     SL &= SSQ\\
     SL &= LLQ\label{eqn:almost_commute}\\
     SL &= LS,\label{eqn:commutes}
 \end{align}
 where \eqref{eqn:commutes} comes from substituting \eqref{eqn:mod_connection} into \eqref{eqn:almost_commute}.\newline

 \noindent\textit{Proving $S=L$:}  Because $S$ and $L$ are symmetric commuting matrices we have that $S$ and $L$ are simultaneously diagonalizable (see Theorem 1.3.12 in \cite{LinAlg_book}).  That is, there exists invertible matrix $H$ such that $S = H^{-1}D_SH$ and $L = H^{-1}D_LH$ 
 for diagonal matrices $D_S$ and $D_L$ whose diagonals contain eigenvalues of $S$ and $L$ respectively.  Let us refer to the diagonal entries of $D_S$ as $D_i(S)$ and the diagonal entries of $D_L$ as $D_i(L)$ for $1\leq i \leq K$. Now note that
 \begin{align}
     SS &= LL\notag\\
     H^{-1}D_S^2H &= H^{-1}D_L^2H\notag\\
     D_S^2 &= D_L^2\label{eqn:equal_squares}.
 \end{align}
 Here  $D_S^2 = D_SD_S$ is the diagonal matrix whose entries are the square of the entries of $D_S$.  Similarly, the elements of $D_L^2$ are the square of the entries of $D_L$.  Then \eqref{eqn:equal_squares} implies that $D_i(S)^2=D_i(L)^2$  and $D_i(S)=\pm D_i(L)$ for all $1\leq i \leq K$.  Because $D_i(S)$ and $D_i(L)$ contain the eigenvalues of $S$ and $L$ which are all positive ($S$ and $L$ are positive definite), the corresponding elements $D_i(S)$ and $D_i(L)$ cannot have different signs and $D_i(S)=D_i(L)$ for all $1 \leq i \leq K.$ This implies that
 \begin{align*}
     D_S &= D_L\\
     H^TD_SH &= H^TD_LH\\
     S&=L.\blacksquare
 \end{align*}

 Theorem \ref{thm:unique_root} tells us that $S$ may, theoretically, be recovered from its row Gramian $G$.  In the next section we explore the properties of $G$ and how it may be estimated from the training data.

\subsection[Estimating G using  a Pair-Wise Contrastive Model]{Estimating $G$ using  a Pair-Wise Contrastive Model}\label{sec:pair-wise_contrast}  Now that we know $G$ is sufficient to recover $S$ in reasonable classification scenarios with uniform prior probabilities, we present a detailed explanation of how $G$ may be estimated.  We begin by analyzing $G$'s basic properties. 
\begin{definition}[Row Gramian of the Collision Matrix]
     If $S$ is a collision matrix, then its row Gramian matrix $G$ is defined by $G = SS^T$.  $G$ is symmetric positive semi-definite and its elements satisfy 
     \begin{equation}   
         G_{i,j} = \mathop{\mathbb{E}}_{\mathbf{x}\sim \mathcal{D}_i}\left [ \mathop{\mathbb{E}}_{\tilde{\mathbf{x}}\sim \mathcal{D}_j} \left [\mathbb{P}(c=\tilde{c}|\mathbf{x},  \tilde{\mathbf{x}}) \right ] \right ]\label{eqn:gramian_element}.         
     \end{equation}
     Additionally, if $\boldsymbol{\pi} = (1/K,1/K,...,1/K)$, then $G$ is doubly stochastic.
 \end{definition}

\noindent We can derive  \eqref{eqn:gramian_element} by expanding the elements of $SS^T$:

 \begin{align}
     G_{i,j} &= \sum_{k=1}^{K}S_{i,k}S_{j,k}\notag\\
     &= \sum_{k=1}^K \mathop{\mathbb{E}}_{\mathbf{x}\sim \mathcal{D}_i} [ \mathbb{P}(c=k|\mathbf{x}) ] \mathop{\mathbb{E}}_{\tilde{\mathbf{x}}\sim \mathcal{D}_j} [ \mathbb{P}(\tilde{c}=k|\tilde{\mathbf{x}}) ]\notag\\
     &= \mathop{\mathbb{E}}_{\mathbf{x}\sim \mathcal{D}_i}\left [ \mathop{\mathbb{E}}_{\tilde{\mathbf{x}}\sim \mathcal{D}_j} \left [\sum_{k=1}^K \mathbb{P}(c=k|\mathbf{x})\mathbb{P}(\tilde{c}=k|\tilde{\mathbf{x}}) \right ] \right ]\notag\\
     &= \mathop{\mathbb{E}}_{\mathbf{x}\sim \mathcal{D}_i}\left [ \mathop{\mathbb{E}}_{\tilde{\mathbf{x}}\sim \mathcal{D}_j} \left [\mathbb{P}(c=\tilde{c}|\mathbf{x},  \tilde{\mathbf{x}}) \right ] \right ].
     \label{eqn:Gramian_element}
 \end{align}
The last equality above takes advantage of the fact that $\mathbf{x}$ and $\tilde{\mathbf{x}}$ are drawn independently from $\mathcal{D}_i$ and $\mathcal{D}_j$ respectively.   Consider the probability on the inside of \eqref{eqn:Gramian_element} which represents the likelihood that two inputs $\mathbf{x}$  and $\tilde{\mathbf{x}}$ will belong to the same class.  This is an important measurement of the similarity between two inputs which we formally define below.
 \begin{definition}[Similarity]\label{def:sim}
    We define the similarity between two points $\mathbf{x}$ and $\tilde{\mathbf{x}}$ as
\begin{equation}\label{eqn:similarity}
    \mathrm{Sim}(\mathbf{x},\tilde{\mathbf{x}}) := \mathop{\mathbb{P}}_{\substack{(\mathbf{x},c)\sim\mathcal{D}\\(\tilde{\mathbf{x}},\tilde{c})\sim\mathcal{D}}}(c=\tilde{c}|\mathbf{x},  \tilde{\mathbf{x}}).
\end{equation}    
\end{definition}
\noindent This similarity score can equivalently be interpreted as follows: for a fixed pair of inputs $(\mathbf{x}, \tilde{\mathbf{x}})$, if we draw the class random variables $c \sim \mathbf{y}(\mathbf{x})$ and $\tilde{c}\sim \mathbf{y}(\tilde{\mathbf{x}})$ independently, the similarity is the probability that both class labels will be equal, $c=\tilde{c}$.  As opposed to other measures of similarity such as a mathematical metric, this similarity score relates more to the semantic information contained in the inputs.  Consider a practical example in healthcare diagnostics: two patients would be considered similar not because they look similar or have similar measurements but rather because their test results suggest that both patients are likely to suffer from the same disease.  We next show how this similarity score can be estimated. \newline\newline
\noindent\textbf{{Pair-Wise Contrastive Model}:} Consider a binary classifier $V:\mathcal{X}\times\mathcal{X}\rightarrow[0,1]$ that takes in two inputs simultaneously and outputs their similarity from Definition \ref{def:sim}.  We call $V$ the \textit{pair-wise contrastive model}.  We first developed this model for the separate task of verifying that modified data points (actionable perturbations) have the desired real-world effect in \cite{TAP}, however this model is also very useful for estimating $G$.  In Section \ref{sec:posterior_estimation} we will show that this model is also useful for estimating class posterior distributions.
 
The first step to training such a model is to create training data suited to its task. To create this \textit{difference training data}, we form every possible pair of inputs from the one-hot labeled dataset $\mathcal{T}$ and label the pairs by whether their class labels match.  Specifically,
\begin{equation}\label{eqn:diff_data}
\mathcal{T}_{\text{diff}}=\{(\mathbf{x}^{(i)}, \mathbf{x}^{(j)}), z^{(i,j)}|\mathbf{x}^{(i)}, \mathbf{x}^{(j)}\in\mathcal{T}\}
\end{equation} where the new label, $z^{(i,j)} = \mathbbm{1}(c^{(i)} = c^{(j)})$, indicates whether the pair belongs to the same class.  Interestingly,  the cardinality of the difference training data $\mathcal{T}_{\text{diff}}$ is $n^2$.  This radically increases the number of training inputs, enabling us to train contrastive models even in settings with very few elements in the initial training data $\mathcal{T}$.  

Although $\mathcal{T}_{\text{diff}}$  encapsulates the relationship we wish to learn (i.e., if a pair of inputs is in same class or different classes), merely training a model to minimize loss on $\mathcal{T}_{\text{diff}}$ does not produce an effective classifier due to the inherent imbalances in the types of pairs $(\mathbf{x}^{(i)},\mathbf{x}^{(j)})$ found in $\mathcal{T}_{\text{diff}}$.  Consider the scenario with $K = 10$ classes and uniform prior probabilities $\boldsymbol{\pi}$: the expected proportion of pairs in $\mathcal{T}_{\text{diff}}$ labeled $1$ (same class)  is only $10\%$, and the constant classifier $V(\mathbf{x},\tilde{\mathbf{x}}) =0$ would achieve $90\%$ accuracy despite providing no useful insight into the relationship between $\mathbf{x}$ and $\tilde{\mathbf{x}}$.  In order to learn an effective pair-wise classifier we must define a learning objective that places equal importance on identifying pairs of inputs from the same or different classes.  To this end we use a modified version of classifier risk and empirical classifier risk for training such a pair-wise contrastive model:
\begin{align}
    R(V) =& \frac{1}{K(K -1)} \sum_{i \neq j}\mathop{\mathbb{E}}_{\mathbf{x}\sim\mathcal{D}_j, \tilde{\mathbf{{x}}}\sim\mathcal{D}_k}[\ell(0,V(\mathbf{x},\tilde{\mathbf{x}}))]+\frac{1}{K} 
 \sum_{k=1}^{K}
\mathop{\mathbb{E}}_{\mathbf{x},\tilde{\mathbf{x}}\sim\mathcal{D}_k}[\ell(1,V(\mathbf{x},\tilde{\mathbf{x}}))]\label{eqn:V_risk}\\
\hat{R}_{\mathcal{T}}(V) =& \frac{1}{K(K -1)} \sum_{i \neq j} \frac{1}{\left(\frac{n}{K}\right)^2} \sum_{\tilde{\mathbf{x}} \in \mathcal{T}_i} \sum_{\mathbf{x} \in \mathcal{T}_j}\ell(0,V(\tilde{\mathbf{x}},{\mathbf{x}}))+\frac{1}{K} 
 \sum_{i=1}^{k} \frac{1}{\left(\frac{n}{K}\right)^2 - n}
\sum_{\substack{\mathbf{x},\tilde{\mathbf{x}} \in \mathcal{T}_i\\ \mathbf{x}\neq\tilde{\mathbf{x}}}}\ell(1,V(\tilde{\mathbf{x}},{\mathbf{x}})).\label{eqn:V_E_risk}
\end{align}
To minimize this risk one may use gradient based optimization methods and sample input pair types from $\mathcal{T}_{\text{diff}}$ according to their proportional importance in \eqref{eqn:V_risk} and  \eqref{eqn:V_E_risk}.  Models trained to minimize $\hat{\mathrm{R}}_\mathcal{T}(V)$ are PAC-learnable as shown in Theorem \ref{thm:PAC}. 

\begin{theorem}\label{thm:PAC}
   Let $V$ be selected from a function class $\mathcal{V}$ and $\ell(\cdot,\cdot)\leq B_\ell$ is a bounded loss function.  Then for any $\delta\in(0,1)$, with probability $(1-\delta)$, the following bound holds.
   \begin{equation}
       \sup_{V\in\mathcal{V}}\left | \mathrm{R}(V)- \hat{\mathrm{R}}_\mathcal{T}(V) \right | =  \mathcal{O}\left ( \left (\frac{K}{\sqrt{n^2-K^2n}} \right )^{1/d} \right ) \label{eqn:PAC}
       {+ \frac{12kB_\ell}{\sqrt{n^2 -K^2n}} \sqrt{\frac{\log(2/\delta)}{2}}}
   \end{equation}
   Here the probability is over the random sampling of $\mathcal{T}$.
\end{theorem}
The generalization gap \eqref{eqn:PAC} provides assurance that, even when training to minimize this new empirical risk definition \eqref{eqn:V_E_risk}, performance should generalize to the broader distribution.    
A detailed proof of Theorem \ref{thm:PAC} is found in \cite{TAP}. {We expect the $\delta$ containing term to be dominated by the other term because $(\frac{K}{\sqrt{n^2-K^2n}})<1$ and $d$ is expected to be much larger than $1$.  Furthermore, we n}ote that {\eqref{eqn:PAC}} differs from typical PAC bounds (Theorem 4.3 in \cite{ml_reference}) which lack the strong dependence on $K$.  This implies that, although the verifier is learnable, it is more sensitive to the number of classes and larger training datasets are necessary in scenarios with large class numbers {and the amount of training data $n$ must increase at a rate greater than $\sqrt{K}$ in order to prevent the denominator from approaching zero}.  In practice, we find that effective pair-wise contrastive models can be created by using the same model architecture for $V$ as one would use for a traditional classifier on the same dataset.  For example, in the case of neural networks, we use the same number of hidden layers and size of hidden layers as we would use for a traditional classifier, only changing the size of the input and output (necessary so that $V:\mathcal{X}\times\mathcal{X}\rightarrow[0,1]$).  We also find that $V$ trains in a similar amount of time to a traditional classifier on the same architecture on the same dataset.  {F}or more information on the pair-wise contrastive model please see \cite{TAP}.

The first step of our process to estimate the collision matrix $S$ will be to train such a pair-wise contrastive classifier $V$ to minimize our newly defined risk $R(V)$.  Once $V$ is obtained, we estimate the entries of $G$ by replacing the expectations in \eqref{eqn:gramian_element} with the empirical averages and replacing $\mathbb{P}(c=\tilde{c}|\mathbf{x},  \tilde{\mathbf{x}})=\mathrm{Sim}(\mathbf{x},\tilde{\mathbf{x}})$ with $V(\mathbf{x},\tilde{\mathbf{x}})$ to arrive at 
\begin{equation}
\hat{G}_{i,j} \approx \frac{1}{|\mathcal{T}_i||\mathcal{T}_j|}\sum_{\mathbf{x} \in \mathcal{T}_i} \sum_{\tilde{\mathbf{x}} \in \mathcal{T}_j} V(\mathbf{x},\tilde{\mathbf{x}}). 
\label{eqn:gram_element_est}
\end{equation}


\subsection[Algorithm for estimating S and Complexity]{Algorithm for estimating $S$ and Complexity}\label{sec:collision_algorithm} Now that we have a method for  estimating of $G$, we may lay out our full process for estimating $S$.  From Theorem \ref{thm:unique_root} we know that, under reasonable assumptions, $S$ should be uniquely defined by $G$, but we still need an algorithm to find $S$.  We propose defining an optimization problem 
\begin{equation}
    \min_{\hat{S}\in\mathcal{S}} \hspace{5pt} || \hat{S} \hat{S}^T - \hat{G} ||_F,
    \label{eqn:optimization}
\end{equation} 
where $\mathcal{S}$ is the set of diagonally dominant stochastic matrices.  To solve \eqref{eqn:optimization} we use gradient descent with an added penalization term to ensure that $\hat{S}$ is stochastic.  The penalty term includes (a) the sum of the absolute difference between the row sums of $\hat{S}$ and $1$ and (b) the sum of  negative values of $\hat{S}$:
\begin{equation}
    \mathrm{arg}\min_{\hat{S}}   || \hat{S} \hat{S}^T - \hat{G} ||_F +\lambda\left (||\hat{S}\mathbf{1}-\mathbf{1}||_1 - \sum_{i=1}^K\sum_{j=1}^K \min\{\hat{S}_{i,j},0\}\right ).\label{eqn:objective}
\end{equation}
Here $\lambda$ is an appropriately sized constant.  

\noindent{\textbf{Estimating $S$ with Non-Uniform Priors:} In the case where the prior distribution $\boldsymbol{\pi}$ is not uniform, we cannot guarantee a unique decomposition of $G$.}  We then use the $K\times K$ identity matrix as our initial guess for $\hat{S}$ during gradient descent.  This biases our search to the possible collision matrix which indicates the least overall uncertainty in the system (When $S=I$ there is no aleatoric uncertainty in $\mathcal{D}$).  This is motivated by the fact that we expect real-world data to exhibit more structure (less aleatoric uncertainty) than random artifacts (such as other matrices satisfying $LL^T = G$).
We now have the three essential components of our method to estimate $S$.

\begin{enumerate}
    \item A method for training the Pair-Wise Contrastive Model $V$.
    \item A process for estimating the row Gramian matrix $G$ using $V$ via \eqref{eqn:gram_element_est}.
    \item A method for recovering $S$ from $G$ by solving \eqref{eqn:objective}.
\end{enumerate}Algorithm \ref{alg:collision_estimation} presents the full process for estimating the collision matrix $S$ from training data $\mathcal{T}$.

\begin{algorithm}[t]
\caption{Collision Matrix Estimation via Pair-Wise Contrast}\label{alg:collision_estimation}
\begin{algorithmic}
\STATE {\bfseries Input:} $\mathcal{T} = \{\mathcal{T}_1,\mathcal{T}_2,\cdots,\mathcal{T}_K\}$ partitioned training data and learning rate $\eta$.
\STATE {\bfseries {Part 1: Estimating the row Gramian matrix $G$}}

\begin{ALC@g}
\STATE \textbf{Step A:} Train pair-wise contrastive model $V$ according to Section \ref{sec:pair-wise_contrast}.

\STATE \textbf{Step B:} Use $V$ to estimate the entries of the row Gramian matrix $G$ using \eqref{eqn:gram_element_est}.
\bindent
\FOR{$i$ between $1$ and $K$} 
\FOR{$j$ between $1$ and $K$} 
\STATE $G_{i,j} \gets \frac{1}{|\mathcal{T}_i|}\sum_{\mathbf{x} \in \mathcal{T}_i} \frac{1}{|\mathcal{T}_j|}\sum_{\tilde{\mathbf{x}} \in \mathcal{T}_j} V(\mathbf{x},\tilde{\mathbf{x}})$
\ENDFOR
\ENDFOR

\eindent
\end{ALC@g}
\STATE {\bfseries {Part 2: Estimating  Collision Matrix $S$}}

\begin{ALC@g}
\STATE \textbf{Step C:} Solve \eqref{eqn:objective} via gradient descent to get the estimated collision matrix.
\bindent
\STATE  $S \gets I$ (or any diagonally dominant matrix)

\WHILE{$\|S S^T - G\|_{F} > \gamma$}
\STATE $\mathbf{a} \gets  {\nabla}_S|| S S^T - G ||_F $
\STATE$\mathbf{b} \gets {\nabla}_S\left ( ||S\mathbf{1}-\mathbf{1}||_1- \sum_{i=1}^K\sum_{j=1}^K \min\{S_{i,j},0\} \right )$
\STATE $S \gets S - \eta (\mathbf{a}+\lambda \mathbf{b})$

\ENDWHILE
\eindent
\end{ALC@g}
\STATE return $S$
\end{algorithmic}
\end{algorithm}

\textbf{Complexity:} We may now evaluate the computational complexity of Algorithm \ref{alg:collision_estimation} which  is divided into two parts: (a) compute the row Gramian $G$ and (b) recover $S$ from $G$.  We will compare the tasks involved in each of these steps (recall $K$ is the number of classes and $n$ is the number of points in the training data):
\begin{itemize}
    \item[a)] We first train the pair-wise classifier $V$, then compute an empirical estimate for each of the $K^2$ elements of $G$ by calculating the average of a series of outputs from $V$.  These empirical estimates require $n^2$ evaluations of $V$, one evaluation for each element of $\mathcal{T}_{\text{diff}}$.  In practice, however, we found it was equally effective to pick a number $m$ of randomly selected elements to use in each average.  This strategy requires only $mK^2$ evaluations of $V$.
    \item[b)] Run a $K^2$ dimensional gradient descent  on optimization problem \eqref{eqn:objective} to find our estimate $\hat{S}$.
\end{itemize}
In practice, the training of $V$ dominates all other tasks.  The PAC-bound from Theorem $2$ shows that $V$ is learnable, but it does not indicate how complex the model $V$ should be nor how long it will take to train.  In all of our experiments, however, we found that we could train an effective $V$ by using similar model architecture as one would use for an effective traditional classifier $M$ on the same data set.  Furthermore, we found that the traditional classifier $M$ and the pair-wise contrastive classifier $V$ require similar amounts of time to train.  This means that training $V$ involves optimizing across all the parameters of $V$ which will generally include more than $K^2$ values and is the largest computational challenge in Algorithm $1$.  Accordingly, we find that Algorithm $1$ can realistically be applied to estimate the collision matrix on any dataset where it is possible to train an effective traditional classifier.  One important note is: if there are $n$ points in the training data $\mathcal{T}$, there are $n^2$ pairs of data points in the difference training data $\mathcal{T}_{\text{diff}}$ (see equation \eqref{eqn:diff_data}).  Thus, our method can be applied even to settings with a limited number of points in the training data.

\section{Estimating Posterior Distributions Using the Collision Matrix}\label{sec:posterior_estimation}

As noted in the introduction, the collision matrix describes the aleatoric uncertainty in the entire data distribution $\mathcal{D}$, however, in practical settings one may also wish to describe the uncertainty (and by extension the potential risks) associated with an individual input $\mathbf{x}$.  The  posterior class probability distribution (Definition \ref{def:true_probs}) fully encapsulating the aleatoric uncertainty in classifying $\mathbf{x}$.  Learning these posterior distributions, however, is a very difficult task due to the type of training data available in practical situations.  In the real-world training data has the form of $\mathcal{T}$ where it is labeled by the single class in which the point was observed (e.g. what disease was ultimately determined to cause the symptoms).  These one-hot labels do not contain information on what other classes were probable candidates making it difficult to train models to estimate posterior distributions and to test the performance of such models once they've been trained.

\subsection{Existing Posterior Estimation Methods}\label{sec:prior_posteriors}  Before we describe the how collision matrix and class collisions may be used to estimate posteriors, we discuss existing posterior estimation methods.  A simple strategy to estimate posteriors is to use the output of an ML classifier trained with traditional empirical risk minimization (ERM) methods.  We refer to such classifiers as $M^{\boldsymbol{\theta}}:\mathcal{X}\rightarrow\mathcal{K}$, where $\mathcal{K}$ is the set of probability distributions over $K$ classes and $\boldsymbol{\theta}$ contains the parameters of the model.  The problem with this method is that traditional ML training focuses achieving a high accuracy through minimizing the risk of the classifier defined as 
$\mathbb{E}_{(\mathbf{x},c)\sim\mathcal{D}}\ell(M^{\boldsymbol{\theta}}(\mathbf{x}),c)$,
where $\ell$ is a loss function designed as a differentiable analog for accuracy.  Note that this risk depends only on the observed class $c$, but a good posterior estimate depends on all the class probabilities $\mathbf{y}(\mathbf{x})$.  This means that high accuracy does not imply that the model is effective as estimating posterior distributions, and classifiers trained for ERM have been shown to attempt to replicate one-hot outputs (be overconfident) \cite{overconfident_base,overconfident_driving,overconfident_bayes,overconfident_mixup,overconfident_variation}.  A more suitable goal would be to minimize the expected deviation from the true posterior
$\mathop{\mathbb{E}}_{(\mathbf{x},c)\sim\mathcal{D}}\ell(M^{\boldsymbol{\theta}}(\mathbf{x}),\mathbf{y}(\mathbf{x}))$,
where $\ell$ is now a loss function measuring the statistical divergence between distributions $M^{\boldsymbol{\theta}}(\mathbf{x})$ and $\mathbf{y}(\mathbf{x})$.  Unfortunately, we cannot train for this directly because we do not have access to any posteriors $\mathbf{y}(\mathbf{x})$.

Previous techniques to address this concern and provide better posterior estimates can be broadly divided into two categories:  post-hoc model augmentations \cite{temp_scale,platt,isotonic} and Bayesian inspired ensemble methods \cite{Bayesian_Networks,mc_dropout}.  
Post-hoc model augmentations focus on modifying the output of a normally trained model to improve its ability to estimate posteriors by improving the \textit{calibration} of the model.  A model is well calibrated if its confidence matches its actual performance, e.g. of all a model's predictions where it puts $80\%$ confidence on the top class, the model should be right $80\%$ of the time.
There are a variety of method for improving calibration such as temperature scaling \cite{temp_scale}, Platt Scaling \cite{platt} and isotonic regression \cite{isotonic}, however, it is important to note that being well calibrated is a necessary but not a sufficient condition for accurately estimating posteriors.  Clearly a model that predicts true posteriors will be well calibrated, but a model that does not accurately estimate posteriors could also be well calibrated.  For example, a constant classifier outputting the prior distribution, $M^{\boldsymbol{\theta}}(\mathbf{x}) = \boldsymbol{\pi}$, is well calibrated but will not provide good posterior estimates as it does not use any of the information in $\mathbf{x}$.  This implies methods for improving calibration may not converge to the true posterior.

Bayesian inspired ensemble methods, on the other hand, seek to find a distribution over the parameters of the model, $\boldsymbol{\theta}\sim\mathcal{H}$, with tools such as Bayesian neural networks \cite{Bayesian_Networks} and MC Dropout \cite{mc_dropout}.  By sampling model parameters from $\mathcal{H}$ one can create an ensemble of classifiers.  Feeding an input $\mathbf{x}$ into all the members of this ensemble allows one to create a distribution over which classes $\mathbf{x}$ is most likely to be classified into.  Note that the posterior $\mathbf{y}(\mathbf{x})$, describes the aleatoric uncertainty inherent to $\mathcal{D}$, and does not describe \textit{epistemic} uncertainty which comes from a lack of knowledge about the best classifier to fit the distribution \cite{AvsE1,AvsE2,AvsE3}.  The distributions created by ensemble methods, however, describe both aleatoric and epistemic uncertainty \cite{label_wise_AandE,AvsE4}, inhibiting them from accurately modeling true posterior distributions.


\subsection{Estimating Posteriors Through Similarity}\label{sec:pos_and_sim}

We propose an innovative method for estimating posterior distributions using the collision matrix laid out in Figure \ref{fig:outline}.  Our method is based on comparing a point of interest $\mathbf{x}$ to many labeled data points from all classes.  By comparing $\mathbf{x}$ to a variety of points across all classes, we are able to learn how well $\mathbf{x}$ fits into each class and avoid the tendency of conventional ML classifiers to focus only on the most likely class.  The metric we use to compare $\mathbf{x}$ to other points is  similarity score from Definition \ref{def:sim}, leading to the following definition of expected similarity to each class. 

\begin{definition}[Expected Similarity Scores]  The expected similarity scores of  $\mathbf{x}$ is a $K$-length vector $\mathbf{q}(\mathbf{x})$ defined by
 \begin{equation}\label{eqn:eprob_vec}
    \mathbf{q}_i(\mathbf{x}) = \mathop{\mathbb{E}}_{\tilde{\mathbf{x}}\sim \mathcal{D}_i}\mathrm{Sim}(\mathbf{x},\tilde{\mathbf{x}}).
\end{equation}   
\end{definition}

To estimate this expected similarity score we may use the same pair-wise comparison model $V$ that we use to estimate $S$.  We will also need a set of comparison points $\{\tilde{\mathbf{x}}^{(i;j)}\}_{j=1}^m$ for $1\leq i\leq K$, where each point $\tilde{\mathbf{x}}^{(i;j)}$ belongs to class $i$.  Before training $V$ we split $\mathcal{T}$ into training and validation sets and we choose the comparison points $\tilde{\mathbf{x}}^{(i;j)}$ from the validation set to avoid bias introduced by running evaluations on data used in training.  This leads to the equation for estimating the expected similarity score:

\begin{equation}\label{eqn:approx_simvec}
    \hat{\mathbf{q}}_i(\mathbf{x}) = \frac{1}{m}\sum_{j=1}^m V(\mathbf{x},\tilde{\mathbf{x}}^{(i;j)}).
\end{equation}

\begin{figure*}[t]
    \centering
    \includegraphics[width = 14cm]{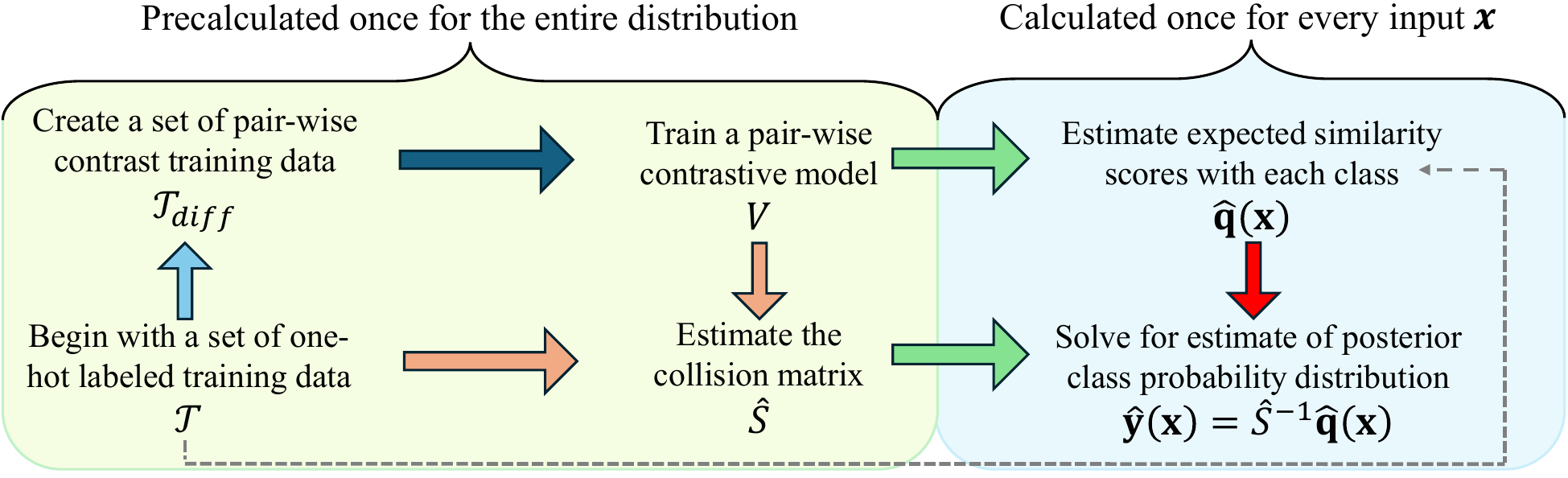}
    \caption{Outline of our \textit{pair-wise contrastive method} for estimating class posterior probability distributions $\mathbf{y}(\mathbf{x})$ from one-hot training data: Starting from one-hot data  $\mathcal{T}$, we create difference training data $\mathcal{T}_{\text{diff}}$ defined in \eqref{eqn:diff_data} consisting of pairs of inputs from either the same or different classes.  We next use $\mathcal{T}_{\text{diff}}$ to train a pair-wise contrastive model $V$ from Section \ref{sec:pair-wise_contrast}. The model $V$ and $\mathcal{T}$ are used to estimate the matrix $S$ (as described in Section \ref{sec:collision_algorithm}) and expected similarity scores $\mathbf{q}(\mathbf{x})$ according to \eqref{eqn:approx_simvec}.  Finally we solve $\hat{S}^{-1} \hat{\mathbf{q}}(\mathbf{x})$ to produce estimated posterior  $\hat{\mathbf{y}}(\mathbf{x})$.}
    \label{fig:outline}
\end{figure*} 


The vector $\mathbf{q}(\mathbf{x})$ indicates how well $\mathbf{x}$ fits in with the points from each class, which intuitively relates to how likely $\mathbf{x}$ is to belong to each class.  The expected similarity scores, however, are mathematically distinct from the true posterior distribution $\mathbf{y}(\mathbf{x})$.  Fortunately, we can connect  the similarity scores $\mathbf{q} (\mathbf{x})$ to the posterior $\mathbf{y}(\mathbf{x})$ by decomposing the probability in \eqref{eqn:eprob_vec}:
\newline
\vspace{-5pt}
\begin{align}
    \mathbf{q}_i(\mathbf{x}) &= \mathop{\mathbb{E}}_{\tilde{\mathbf{x}}\sim \mathcal{D}_i}\mathrm{Sim}(\mathbf{x},\tilde{\mathbf{x}})\notag\\
    &= \mathop{\mathbb{E}}_{\tilde{\mathbf{x}}\sim \mathcal{D}_i}\sum_{j=1}^K\mathbb{P}(\tilde{c}=c=j|\tilde{\mathbf{x}},\mathbf{x})\\
    &= \mathop{\mathbb{E}}_{\tilde{\mathbf{x}}\sim \mathcal{D}_i}\sum_{j=1}^K\mathbb{P}(c=j|\mathbf{x})\mathbb{P}(\tilde{c}=j|\tilde{\mathbf{x}})\label{eqn:indepencence}\\
    &= \sum_{j=1}^K\mathop{\mathbb{E}}_{\tilde{\mathbf{x}}\sim \mathcal{D}_i}\mathbb{P}(\tilde{c}=j|\tilde{\mathbf{x}})\mathbf{y}_j(\mathbf{x})\\
    &= \sum_{j=1}^KS_{i,j}\mathbf{y}_j(\mathbf{x})\label{eqn:systemofeq},
\end{align}
where \eqref{eqn:indepencence} follows from the fact that $(\mathbf{\tilde{\mathbf{x}}},\tilde{c})$ is selected independently from $(\mathbf{x},c)$. Observe that \eqref{eqn:systemofeq} provides us with a linear system of equations connecting the posterior $\mathbf{y}(\mathbf{x})$ with the expected similarity scores $\mathbf{q}(\mathbf{x})$ and the collision matrix $S$. This is the key insight allowing us to use the collision matrix to estimate posterior distributions. We summarize this  result in  Theorem \ref{thm:decomp}. \newline\newline
\begin{theorem}\label{thm:decomp}
    The posterior distribution $\mathbf{y}(\mathbf{x})$, the expected similarity scores $\mathbf{q}(\mathbf{x})$ and the collision matrix $S$ are related by the linear equation
    \begin{equation}\label{eqn:collision}
        \mathbf{q}(\mathbf{x}) = S \mathbf{y}(\mathbf{x}).
    \end{equation}
    When $S$ is non-singular (as guaranteed when $S$ is strictly diagonally dominant) we may write \begin{equation}\label{eqn:linear}
        \mathbf{y}(\mathbf{x}) = S^{-1} \mathbf{q}(\mathbf{x}).
    \end{equation}
\end{theorem}
Equation \eqref{eqn:linear} provides us with simple and efficient way to transform an estimate of the similarity scores $\mathbf{q}(\mathbf{x})$ to and estimate of the posterior distribution $\mathbf{y}(\mathbf{x})$.  Note also that the key components to this method, $\mathbf{q}(\mathbf{x})$ and $S$, may both be estimated through the use of a single tool, the pair-wise comparison model $V$.  This leads to a straightforward process described the the following steps also listed in Algorithm \ref{alg:full}.

\begin{algorithm}[t]
\caption{\textit{Estimating posterior $\mathbf{y}(\mathbf{x})$}}\label{alg:full}
\begin{algorithmic}
\STATE Steps computed \textbf{once} for the distribution.
\vspace{0.05cm}
\hrule
\vspace{0.05cm}
\STATE $V \gets$ Train the pair-wise contrastive model according to the  method in Section \ref{sec:pair-wise_contrast}.

\STATE $\hat{S}\gets$ Estimate the collision matrix using Agorithm \ref{alg:collision_estimation}.
\vspace{0.05cm}
\hrule
\vspace{0.05cm}
\STATE Steps computed for \textbf{each} input $\mathbf{x}$
\vspace{0.05cm}
\hrule
\vspace{0.05cm}
\STATE $\hat{\mathbf{q}}(\mathbf{x}) \gets$ Create an estimate of the expected similarity scores ${\mathbf{q}}(\mathbf{x})$  using \eqref{eqn:approx_simvec}.

\STATE $\hat{\mathbf{y}}(\mathbf{x}) \gets \hat{S}^{-1} \hat{\mathbf{q}}(\mathbf{x})$ Solve \eqref{eqn:linear} to estimate the posterior.
\end{algorithmic}
\end{algorithm}

\begin{enumerate}
    \item Train a pair-wise contrastive model $V$ as described in Section \ref{sec:pair-wise_contrast}.
    \item Use $V$ to estimate the collision matrix as described in Section \ref{sec:collision_algorithm}.
    \item Use the pair-wise contrastive model to compare a new input to one-hot labeled inputs from each class to estimate the expected similarity to each class $\mathbf{q}(\mathbf{x})$.
    \item Multiply the vector of expected similarities to the inverse of the collision matrix, $S^{-1} \mathbf{q}(\mathbf{x})$, to obtain an estimate of the posterior probability distribution.
\end{enumerate}

\noindent{\textbf{Error Bounds:} The error in estimating posterior probability distribution with Algorithm \ref{alg:full} will, of course, depend on the quality of the estimates of the collision matrix $\hat{S}$ and the expected similarity score $\hat{\mathbf{q}}(\mathbf{x})$. These estimates will in turn depend on how well the pair-wise contrastive model learns the data distribution, the number of comparison points used to estimate the collision matrix via Algorithm \ref{alg:collision_estimation} ($|\mathcal{T}_i|$, $1\leq i\leq K$), and the number of comparison points used to estimate the expected similarity scores via \eqref {eqn:approx_simvec} (denoted by $m$). The unique properties of the collision matrix in a reasonable classification scenario (as described in Section \ref{sec:recoverablility}), however, limit the amount to which these errors can propagate into the final estimate $\hat{\mathbf{y}}(\mathbf{x})$ as described in Lemma \ref{lemma:error_bound} below.
 \begin{lemma}\label{lemma:error_bound}
    The estimate $\hat{\mathbf{y}}(\mathbf{x})$ of the true posterior probability distribution ${\mathbf{y}}(\mathbf{x})$ found using Algorithm \ref{alg:full} when $S$ is diagonally dominant satisfies
    \begin{equation}\label{eqn:posterior_error}
        ||\hat{\mathbf{y}}(\mathbf{x})-{\mathbf{y}}(\mathbf{x})||_\infty \leq 2 \left ( \frac{1+\epsilon}{1-\epsilon}\right ) \left ( ||\hat{S}-S||_\infty  +\frac{||\hat{\mathbf{q}}(\mathbf{x})-{\mathbf{q}}(\mathbf{x})||_\infty}{||{\mathbf{q}}(\mathbf{x})||_\infty} \right ),
    \end{equation}
    where
    \begin{align}
        \epsilon := \max_{i}\frac{\sum_{k\neq i}|S_{i,k}|}{|S_{i,i}|}<1\label{eqn:dominance_factor}
    \end{align}
    is the \textit{dominance factor} which grows closer to zero as $S$ grows more diagonally dominant.
\end{lemma}
\noindent This guarantees the element-wise error in our estimate of the posterior distribution, $||\hat{\mathbf{y}}(\mathbf{x})-{\mathbf{y}}(\mathbf{x})||_\infty, $ is bounded by the element-wise error in our estimate of the collision matrix, $||\hat{S}-S||_\infty$, and the relative error in our estimate of the expected similarity score, ${||\hat{\mathbf{q}}(\mathbf{x})-{\mathbf{q}}(\mathbf{x})||_\infty}/{||{\mathbf{q}}(\mathbf{x})||_\infty}$ as long as $S$ is strongly diagonally dominant, i.e. $\epsilon$ not close to $1$.  The proof of Lemma \ref{lemma:error_bound} is presented in Appendix \ref{app:posterior_error}.}

\section{Experiments}\label{sec:experiments}

In this section, we present experimental results justifying our method for estimating the collision matrix  using our two-step method introduced in Section \ref{sec:G_estimation} and our process for estimating class posterior probability distributions in Sections \ref{sec:posterior_estimation}.  We test these methods on both synthetic data, where the true posteriors and collision matrices are known, and on real-world datasets, where we are able to provide useful insights.
Details on model architectures, training procedures \& hyper parameters are included in Appendix \ref{app:details_parameters}.  Our code is also available online\footnote{\href{https://github.com/JesseFriedbaum/UncertaintyViaCollisions}{https://github.com/JesseFriedbaum/UncertaintyViaCollisions}}.

\subsection{Estimating the Collision Matrix}\label{sec:collision_matrix_experiments}
The experimental results in this section illustrate (a) our two-step method implemented in Algorithm \ref{alg:collision_estimation} effectively estimates $S$, outperforming baselines, and (b) that the collision matrix provides useful insights into data not supplied by other UQ methods.  For part (a), to show that we can estimate $S$ effectively we create a variety of synthetic Gaussian mixture datasets where we can calculate true posterior distributions and the ground truth $S$ matrix.  We then estimate $S$ from sampled training data using baseline approaches and our Gramian based method.  We compare the respective estimates from each method to the true $S$ and show an advantage for the Gramian based method. For part (b), to show that $S$ provides useful insights, we estimate $S$ on a variety of real-world datasets.  We first show that our method creates a reasonable collision matrix estimate on image classification data, where we have a high degree of intuition about the features of the true collision matrix.  We next use $S$ to calculate the precision/recall values of the PBC (Probabilistic Bayes Classifier from Definition \ref{def:pbc}) on a variety of tabular datasets form different fields; this allows us to infer what tasks may be accomplished with the data in these datasets and what tasks are not possible with this data.

\begin{figure}[t]
\begin{center}
\includegraphics[width = 14cm]{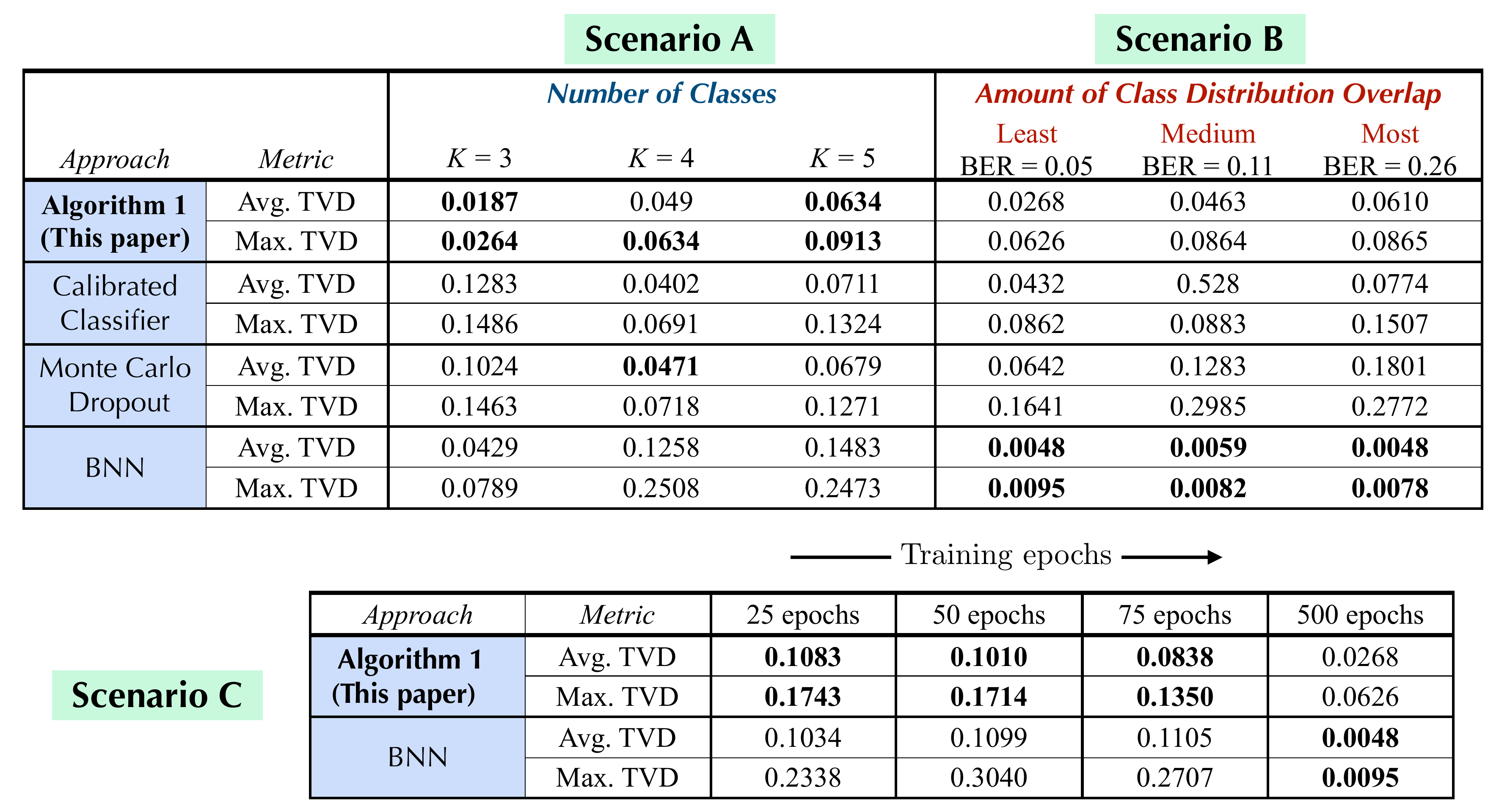}
\end{center}

\caption{Results on Estimating $S$ for Synthetic datasets: comparison of Gramian based approach (Algorithm $1$; this paper) vs. three Baseline approaches-- (1) calibrated classifier; (2) MC dropout and (3) Bayesian neural network (BNN) ensemble. Algorithm $1$ outperforms the baselines in all cases in Scenario A,  outperforms all but the BNN in Scenario B.  We note, BNN only outperforms Algorithm 1 after very long training times (Scenario C).}
\label{img:synthetic-data-results}

\end{figure}

\begin{figure}[t]
\begin{center}
\includegraphics[width = 8cm]{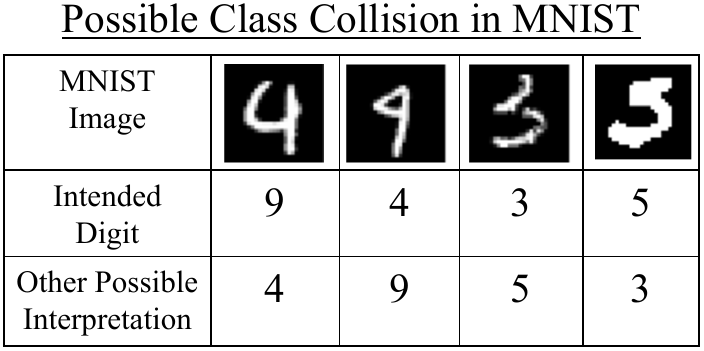}
\end{center}

\caption{Four examples of images in MNIST that could result in class collisions: if these exact same images here drawn on a different occasion it is possible (even likely) that writer intended to draw a different number than on the first occasion.}
\label{fig:mnist_examples}

\end{figure}

\noindent{\textbf{Synthetic Data and Estimating $S$}:} Our synthetic data is generated using Gaussian mixtures: the feature vectors from each class are drawn from a unique multivariate normal distribution: $\mathcal{D}_k = \mathcal{N}(\boldsymbol{\mu}_k, \Sigma)$. We sample training data from these distribution to use in constructing estimates $\hat{S}$ of the collision matrix. As described in Section \ref{sec:G_estimation}, for baseline estimates $\hat{S}$ we solve \ref{eqn:naive_approx} using approximations of the posterior distribution $\mathbf{y}(\mathbf{x}^{(i)})$ from 1) a calibrated classifier, 2) an ensemble model using Monte Carlo (MC) Dropout and 4) an Bayesian Neural Network (BNN) ensemble (see details of these models including a comments on computational complexity in the Appendix \ref{app:synth_models}).  We compare these baselines with our two-step Gramian based Algorithm 1.
In the Gaussian mixture setting we have access to the true collision matrix $S$ which we compare to our estimates $\hat{S}$  the average and maximum \textit{total variation distance} (TVD) between the rows of $\hat{S}$ and the corresponding columns of $S$ (recall that each row of $S$ represents a probability distribution).

\noindent\textit{{Scenario A (impact of varying number of classes)}}. We first report the results for relatively simple $d=4$ dimensional data, where we increase the classification complexity by increasing $K$, the number of classes (Gaussian distribution in the mixture).  Details of the configuration are found in the Appendix \ref{app:synth_details}.  We present our results in Figure \ref{img:synthetic-data-results} where the Gramian based method is the best overall method having lowest maximum TVD in all cases.
\newline
\textit{{Scenario B (impact of class overlap heterogeneity)}}.  For these tests we chose a more complicated setting with $d=20$ dimensional data in $K=5$ classes.  We then varied the amount of uncertainty in the system by moving the datasets closer or further apart to change the amount of overlap between distributions.  We report the results of these experiments in Figure \ref{img:synthetic-data-results} where we also list the BER for each configuration to demonstrate the shift in uncertainty.  Our results show that the Gramian method consistently outperforms all baselines except placing behind the BNN that performs extremely well in this high-dimension setting.  We also note its comparative advantage is larger in the lower overlap settings.\newline
{{\textit{Scenario C (training costs})}}.  Although the BNN performed remarkably well in Scenario B we note that it required many more training epochs to achieve better accuracy than our Algorithm $1$.  Considering that classifier training is the primary computational cost of both Algorithm 1 and baseline methods (see Section \ref{sec:collision_algorithm}} and Appendix \ref{app:synth_models}), this means the Gramian based method has considerable computational benefits.  We present the accuracy of the BNN baseline and the Gramian method at different points in their training in Figure \ref{img:synthetic-data-results}.  The Gramian method is the most well rounded method, performing best on simpler classification problems (less overlap between distributions) and being more efficient to compute on complicated distributions (more overlap between distributions).

\begin{figure}[t]
\begin{center}
\includegraphics[width = 16cm]{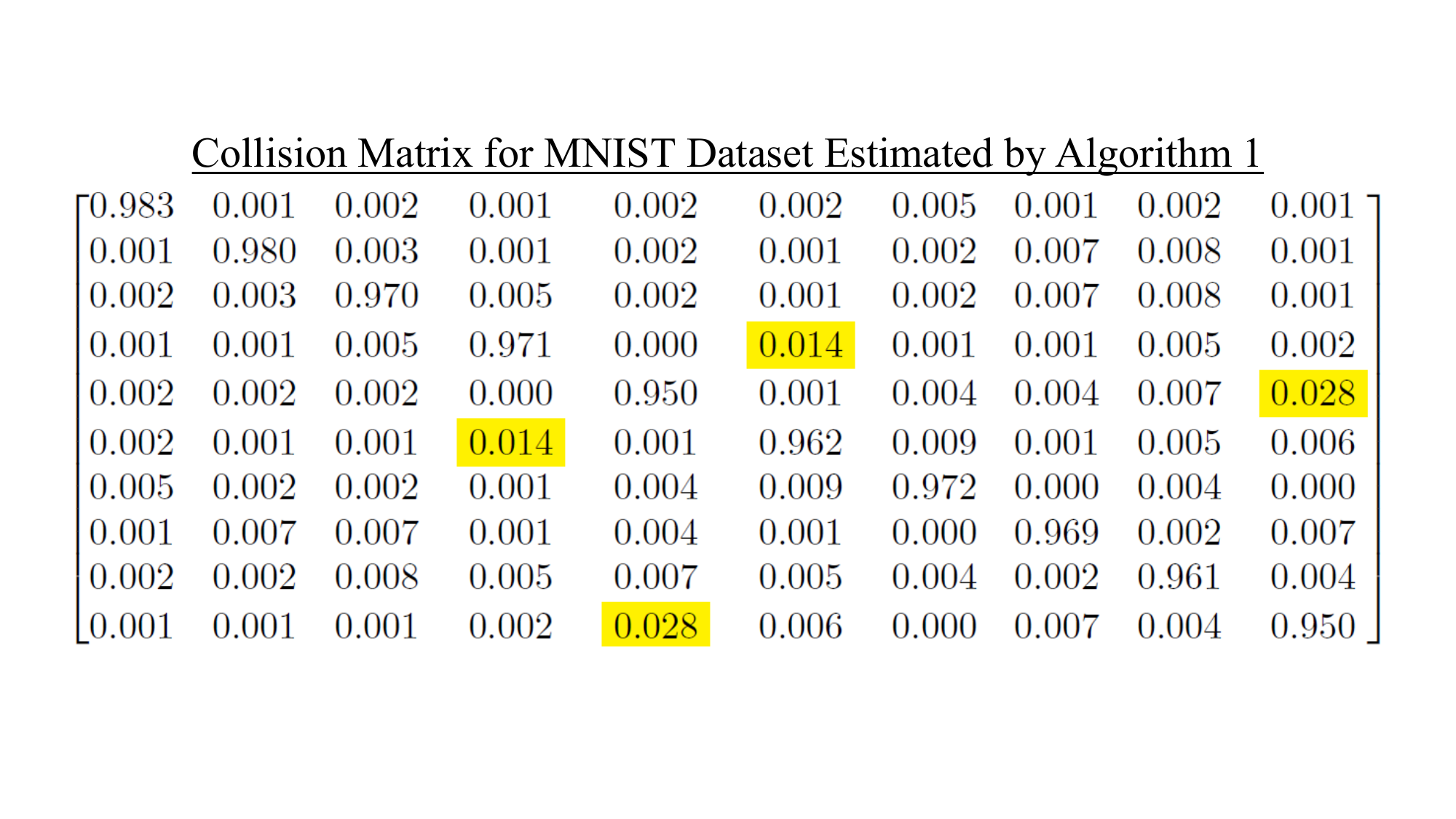}
\end{center}
\caption{Collision matrix $S$ for the MNIST \cite{mnist} dataset found using Algorithm \ref{alg:collision_estimation}.  The largest off diagonal entries are highlighted representing the class collisions between the digits $4$ and $9$ and also between digits $3$ and $5$.  This indicates that there is a small amount of inherent uncertainty in these pairs of digits: it is not always possible to tell what digit the writer intended to make based off of what they actually drew.}
\label{fig:mnist_S}
\end{figure}

\noindent{\textbf{Real-World Data Validation for Estimating $S$}:} We also validate Algorithm \ref{alg:collision_estimation} by estimating the collision matrix on real-world data.  We do not have access to true collision matrix for these real-world datasets, but we may select a dataset for which there is a clear intuition about the form of the collision matrix. Accordingly, we use Algorithm \ref{alg:collision_estimation} to estimate the collision matrix of the MNIST dataset \cite{mnist}  (seen in Figure \ref{fig:mnist_S}).  This image classification dataset contains black and white scans of handwritten digits where the label corresponds to the number that the writer intended to draw.  Many image classification datasets have no aleatoric uncertainty and no possibility for class collision (subject of an image will be the same each time the image is viewed) and are not good candidates for estimating the collision matrix $S$.  MNIST, however, does allow for collisions because the label represents the number the writer intended to draw (the writer's intent) and it is possible that the same drawing could be made multiple times even though the writer's intended to draw distinct numbers.  For examples of this see Figure \ref{fig:mnist_examples}, which contains $4$ images from the MNIST dataset that could reasonably be interpreted as multiple digits.

Our estimate of the collision matrix $S$ is found in Figure \ref{fig:mnist_S}.  Most off-diagonal entries were very small (less than $0.01$) indicating very low likelihood of class collisions and low uncertainty.  For the digit pairs $4$ \& $9$ as well as $3$ \& $5$, we found collision probabilities above $0.01$, indicating these digit pairs had the greatest inherent uncertainty.  Specifically, the fact that $S_{4,9}$ = 0.028 indicates that, if the same exact hand drawn digit that was originally intended to be a $4$ is seen again, there is a $2.8\%$ chance that in the second observation it was meant to be a $9$.  These particular digit pairs are easily confused in the handwriting of some individuals (see Figure \ref{fig:mnist_examples}), and the fact that our estimated $S$ predicted the highest uncertainty for these pairs indicates that Algorithm \ref{alg:collision_estimation} is effective.\newpage

\noindent{\textbf{Real-World Tabular Data and Insights from $S$}:} To show that $S$ can produce useful information on real-world data, we use four tabular datasets containing the clear possibility of class collision (aleatoric uncertainty). 
\begin{enumerate}
    \item \textit{Adult Income dataset} \cite{adult_DS}. Task: predict income bracket using demographic information.
    \item \textit{Law School Success dataset} \cite{law_DS}. Task: predict if a student will pass the BAR examination (legal accreditation) from demographic \& academic records.
    \item \textit{Diabetes Prediction dataset} \cite{diabetes_DS}. Task: predict if an individual is diabetic using information on health conditions and habits
    \item \textit{German Credit dataset} \cite{germancredit_DS} Task: predict credit risk using loan application and demographic data
\end{enumerate}

\begin{figure}[!t]
\begin{center}
\includegraphics[width = \textwidth]{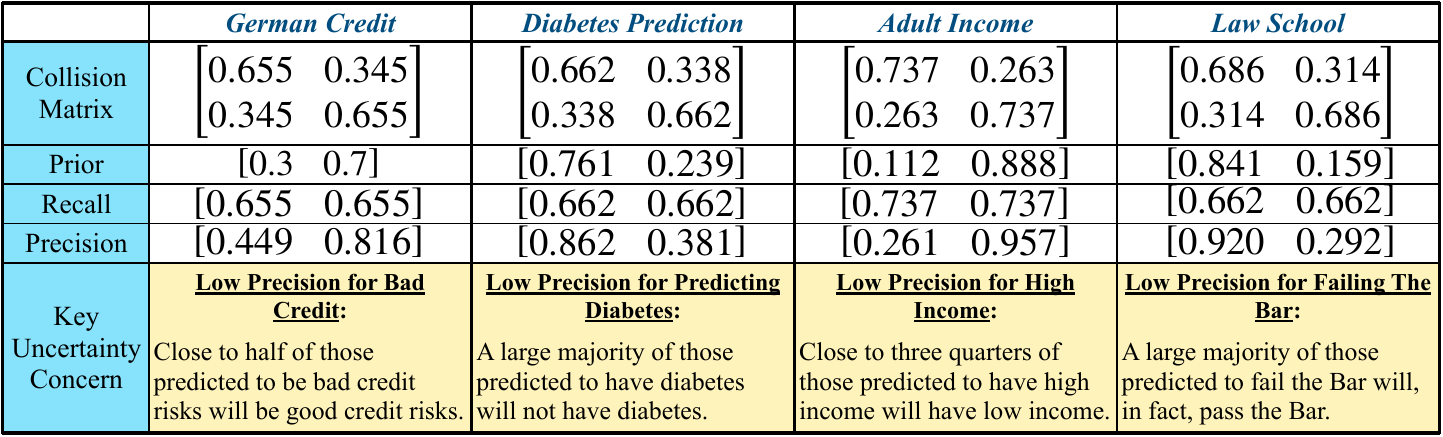}
\end{center}

\caption{\small Results and Insights provided by $S$ for four Real-world datasets: We report the estimated collision matrix and use it to compute the recall and precision of the probabilistic Bayes classifier on these datasets.  We highlight the prediction uncertainties inherent (and therefore unavoidable) in these classification problems.}
\label{fig:real_comparison}

\end{figure}

We estimate the collision matrix using Algorithm $1$ for all datasets, and then find the precision and recall for the probabilistic Bayes classifier (PBC) as described in Section \ref{sec:BER} and Figure \ref{fig:confusion}.  We present our results in Figure \ref{fig:real_comparison}.  Unlike single value uncertainty measures, such as the BER, the collision matrix informs us exactly what types of classification problems are inevitable in the data.  For example, the uncertainty and class imbalances in the adult income dataset will cause low precision for predicting high income such that a most of those predicted to have high income by the PBC will not have high income.  These insights can have important practical implications.  For example, we found very low precision for diabetes (three quarters of those predicted to have diabetes will not have it), but we had much better recall with $67\%$ of individuals with diabetes being correctly classified.  These values suggest: the information in this dataset is useful for identifying individuals who are at high risk for diabetes and should undergo further testing (good recall), but the information in this dataset is not sufficient to make trustworthy diagnoses (low precision).  We emphasize that the collision matrix is a property of the data distribution, so training more sophisticated models on this dataset can not lead to high precision diagnoses, rather richer data (e.g., more informative features) could be necessary before a classifier can achieve high precision.  On the other hand, if a classifier on this dataset achieves recall for diabetes well below $67\%$, we know that it is possible to achieve better recall, suggesting we should adjust the classifier architecture or training procedure.  This level of detail about what kind of classification performance is possible from a dataset is a unique benefit of the fine-grained nature of $S$.

\subsection{Estimating Class Posterior Probability Distributions} 

We also validate our methods for estimating class posterior distributions on synthetic datasets drawn from Gaussian Mixtures and the real-world MNIST dataset.

\begin{figure*}[t]
    \centering
    \includegraphics[width = \textwidth]{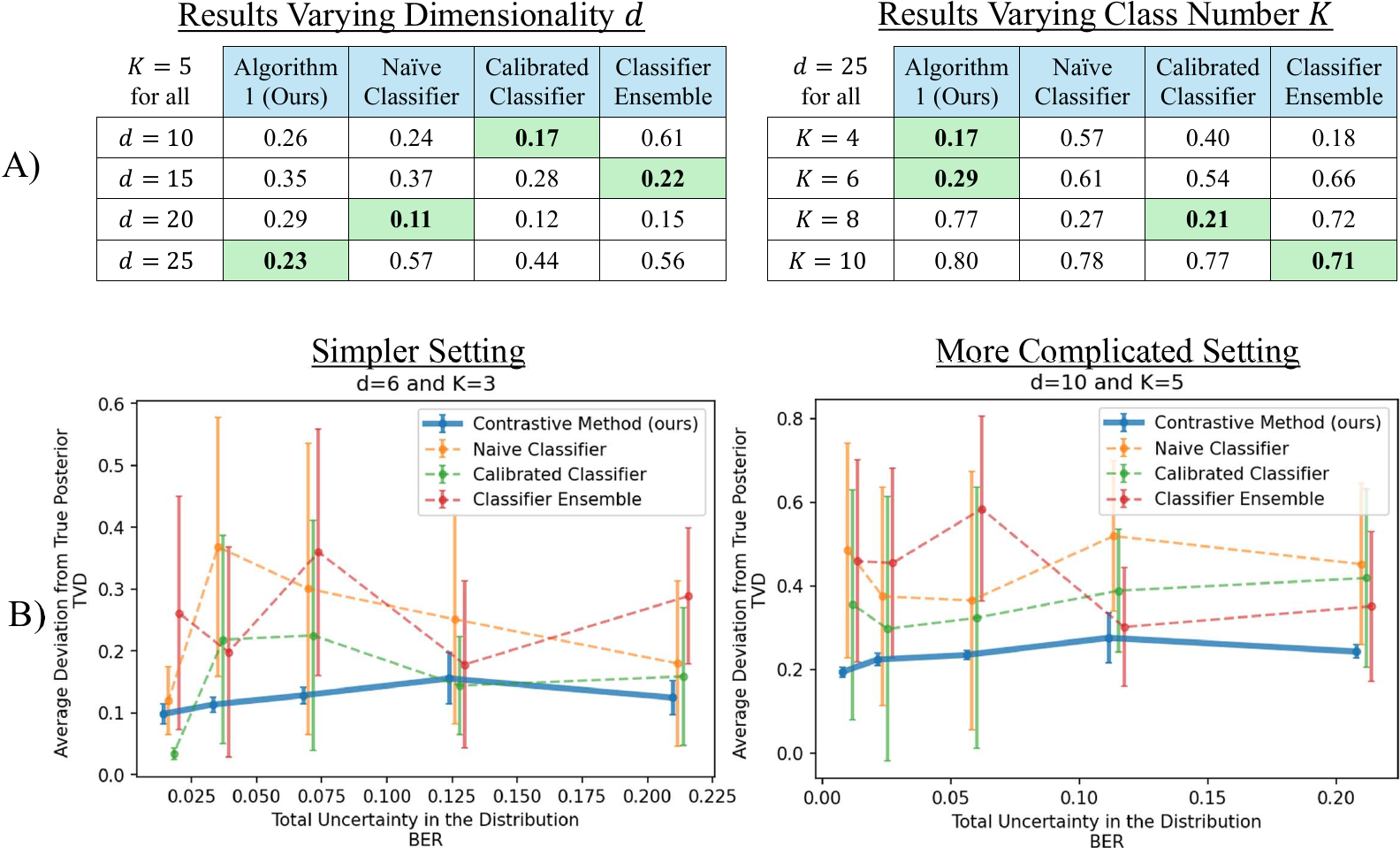}
    \caption{A: Tables containing average Total Variation Distance (TVD) from the true posteriors  to the estimated posteriors using various methods.  Each result represents sampling training data from the distribution \textit{one} time and running all algorithms.  B:  Graphs of the average TVD from the true posteriors  to the estimated posteriors using various methods while varying the total uncertainty in the distribution (measured by BER).  We perform \textit{multiple} tests for each data distribution (resampling training data and rerunning all algorithms).  The  standard deviation of the average TVD from each test is plotted in the error bars.  Our Collision matrix based method exhibits smaller average TVD and far lower variance in estimations.}
    \label{fig:results}
    \vspace{-0.25cm}
\end{figure*} 

\noindent \textbf{Results for Synthetic datasets}. For these tests our Gaussian mixtures consisted of $K$ Gaussian with means distributed  equidistant along a line segment oriented in a randomly selected direction.  We fix the covariances of the distributions so that we may vary the overlap of the distributions---and by extension the aleatoric uncertainty of the distribution---by changing the length of the line segment.  We compare our method against the three baselines from Section \ref{sec:prior_posteriors}: an unmodified (naive) classifier, a classifier calibrated through temperature scaling \cite{temp_scale} and an ensemble of MC-dropout classifiers \cite{mc_dropout}.  We evaluate effectiveness by measuring the average Total Variation Distance (TVD) to the true posterior while varying uncertainty, dimensionality $d$ and class number $K$.  The tables in Figure \ref{fig:results}A, shows which techniques perform best when varying model parameters, where each technique occasionally performs best in at least one test.  To make sense of this we execute many runs of the same experiment (resampling data from distributions with the same parameters) plotted in Figure \ref{fig:results}B, observing that the baseline methods exhibit large variances in performance across runs (explaining why seemingly random techniques would perform best in part A).  We note, however, that our contrastive method exhibits much lower variance and generally lower average TVD.  This indicates our method far more consistent and generally more accurate: clearly the best option.

\begin{figure}[t]
    \centering
    \includegraphics[width = 0.75\textwidth]{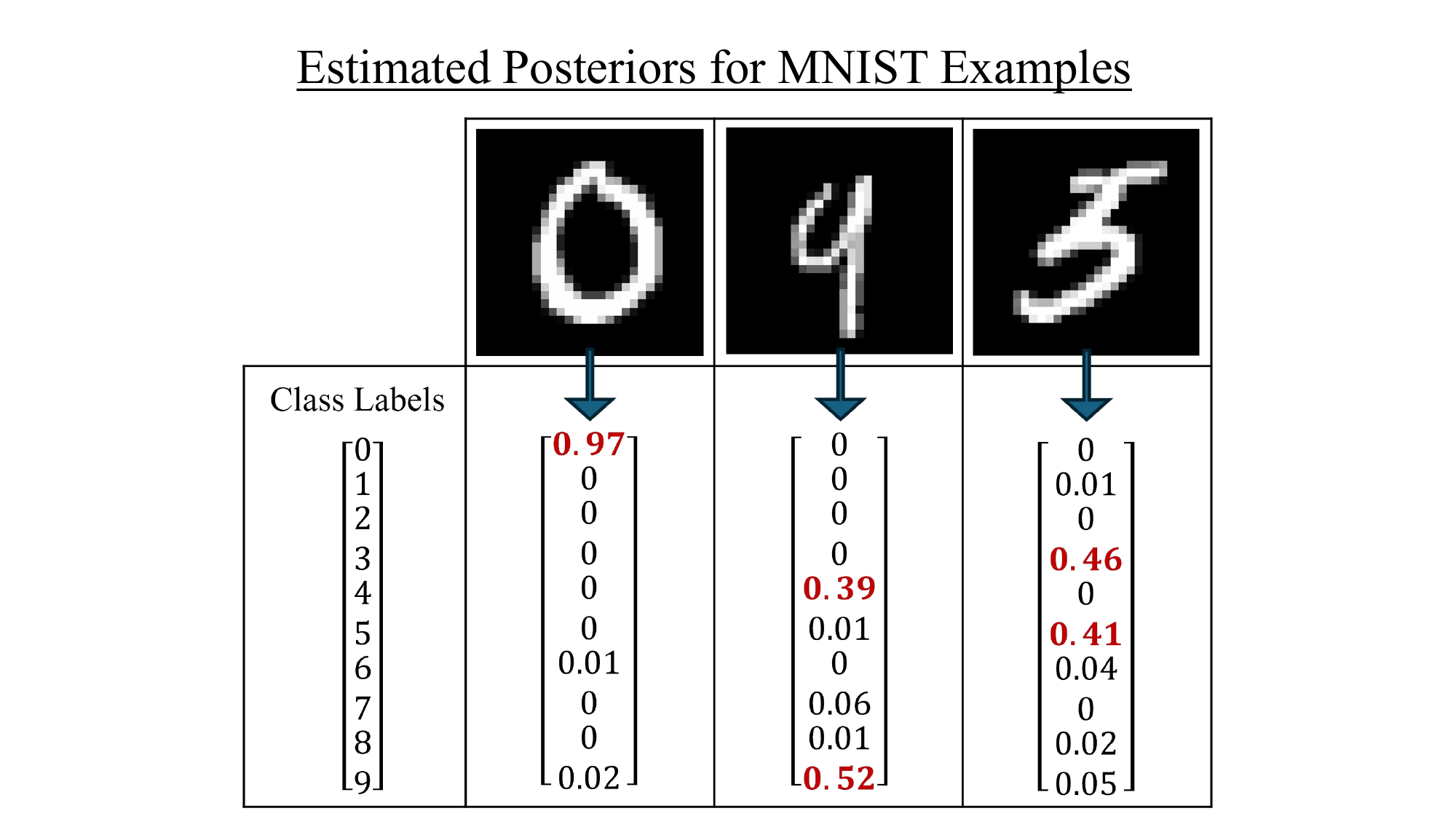}
    \caption{Estimates of the posterior class probability distributions of various MNIST images.  The unambiguous ``$0$'' has a nearly one-hot posterior, whereas  the other two ambiguous images have posteriors that split the probability across the two most plausible classes.}
    \label{fig:mnist_posteriors}
    \vspace{-0.25cm}
\end{figure} 

\noindent \textbf{Insights on Real-world data:} Testing posterior estimates on real-world data is challenging because exact posteriors are never available for real-world datasets containing aleatoric uncertainty (the posteriors are not one-hot).  Fortunately, the same characteristics that made the MNIST \cite{mnist} dataset a useful test case for estimating collision matrices also makes it appropriate for validating our posterior estimation technique from Algorithm \ref{alg:full}. Namely that MNIST contains images are ambiguous and could reasonably represent multiple digits, and we can visually identify which digits the character is most likely to represent.  This should correspond to the largest values in the posterior. In Figure \ref{fig:mnist_posteriors} we present three characters from MNIST and their posteriors as estimated by Algorithm \ref{alg:full}.  We note that for the unambiguous ``$0$''  the posterior is nearly one-hot, but for the ambiguous characters the posterior puts nearly equal weight on the main two possibilities, indicating the our method is producing reasonable posterior estimates on this dataset.

\section{Conclusion} 
We introduced the \textit{Collision Matrix} as a new fine-grained approach for capturing aleatoric uncertainty in classification problems.  Collision matrix provides a uniquely detailed description of uncertainty, measuring the difficulty in distinguishing between pairs of classes and providing unique insights for classification tasks.  We developed the fundamental properties of $S$; and leveraged those to devise a novel pair-wise contrastive approach for estimating $S$. Furthermore, we validated the effectiveness of our estimation algorithm as well as showed the kind of insights that can be drawn from $S$ on several real-world datasets.

We also highlighted one important application of the collision matrix: estimating posterior class probability distributions.  This application allows us to describe uncertainty on the individual input level and better evaluate the risk involved with each decision.  We were able to do this by showing that the posterior distributions can be written as the product of the collision matrix and expected similarity score that is easily estimated using one of the same tools we use to estimate the collision matrix.

\newpage

\bibliographystyle
{IEEEtran}
\bibliography{references}

\newpage
\section{Appendix}


\subsection{Additional Proofs \& Theoretical Results}\label{app:proofs}
\subsubsection{Proof of Proposition \ref{thm:collision_properties}}\label{app:s_properties}

\begin{duplicateprop}[Properties of $S$]
    The entries of the collision matrix $S$ are defined by
    \begin{equation*}
        S_{i,j} = \mathop{\mathbb{E}}_{\mathbf{x}\sim \mathcal{D}_{i}}\mathbb{P}(C=j|\mathbf{x}) = \boldsymbol{\pi}_{j}\int_{\mathcal{X}}\frac{f_{{i}}(\mathbf{x})f_{{j}}(\mathbf{x})}{\sum_{k=1}^K \boldsymbol{\pi}_{k} f_{{k}}(\mathbf{x})}d\mathbf{x}.\tag{\ref{eqn:element_expanded}}
    \end{equation*}
    $S$ is a row stochastic matrix. In the special case with uniform class priors, i.e., $\boldsymbol{\pi} = \left(\frac{1}{K},\frac{1}{K},...,\frac{1}{K}\right)$, 
    $S$ is a symmetric matrix and, therefore, doubly stochastic.
\end{duplicateprop}

   We begin by showing that the closed form expression \eqref{eqn:element_expanded} is equal to the description of the elements of $S$ from Definition \ref{def:confusability_matrix}. We note that we define the space $\mathcal{X}$ containing the feature vectors as containing only possible feature, i.e.,  for all $\mathbf{x}\in\mathcal{T}$ there exists at least one $k\in\{1,2,...,K\}$ such that $f_{k}(\mathbf{x}) \neq 0$.  This is significant as it means that we need not worry about the term $\sum_{k=1}^K \boldsymbol{\pi}_{k} f_{{k}}(\mathbf{x})$, found in the denominator of several fractions, ever being zero. 
   
   Let $\mathcal{B}_\epsilon(\mathbf{x})$ be an $\epsilon$ ball in $\mathbb{R}^d$ centered around $\mathbf{x}$.
\begin{align}
    \mathbb{P}(C=j|X =\mathbf{x}) &= \lim_{\epsilon \rightarrow 0 } \mathbb{P}(C=j|X \in \mathcal{B}_\epsilon(\mathbf{x}))\notag\\
    &=\lim_{\epsilon \rightarrow 0 }\frac{\mathbb{P}(X \in \mathcal{B}_\epsilon(\mathbf{x})|C=j)\mathbb{P}(C=j)}{\mathbb{P}(X\in\mathcal{B}_\epsilon(\mathbf{x}))}\notag\\
    &=\lim_{\epsilon \rightarrow 0 }\frac{\boldsymbol{\pi}_{j}\int_{\mathcal{B}_\epsilon(\mathbf{x})} f_{{j}}(\mathbf{z})d\mathbf{z}}{\int_{\mathcal{B}_\epsilon(\mathbf{x})} \sum_{k=1}^K \boldsymbol{\pi}_{k} f_{{k}}(\mathbf{z})d\mathbf{z}}\notag\\
    &=\frac{\boldsymbol{\pi}_{j} f_{{j}}(\mathbf{x})}{\sum_{k=1}^K \boldsymbol{\pi}_{k} f_{{k}}(\mathbf{x})}\label{eqn:prob_expanded}
\end{align}
Applying an expectation to  \eqref{eqn:prob_expanded} we arrive at \eqref{eqn:element_expanded}:
\begin{equation*}
    \mathop{\mathbb{E}}_{\mathbf{x}\sim \mathcal{D}_{i}}\mathbb{P}(C=j|X =\mathbf{x})=\boldsymbol{\pi}_{j}\int_{\mathcal{X}}\frac{f_{{i}}(\mathbf{x})f_{{j}}(\mathbf{x})}{\sum_{k=1}^K \boldsymbol{\pi}_{k} f_{{k}}(\mathbf{x})}d\mathbf{x}.
\end{equation*}

We now show that $S$ is row stochastic:  the elements of $S$ are non-negative and each row sums to $1$.  Clearly each element of $S$ given by  $\mathop{\mathbb{E}}_{\mathbf{x}\sim \mathcal{D}_{i}}\mathbb{P}(C=j|X =\mathbf{x})$ is non-negative, so we need only show that each row of $S$ sum to $1$.  For ease of notation we will refer to the $k^{\text{th}}$ row of $S$ as
 \begin{equation}\label{eqn:ccv_def}
    \mathbf{s}^T(k) = 
    \begin{pmatrix}
        \mathop{\mathbb{E}}_{\mathbf{x}\sim \mathcal{D}_k}\mathbb{P}(C=1|X =\mathbf{x})&
        \cdots&
        \mathop{\mathbb{E}}_{\mathbf{x}\sim \mathcal{D}_k}\mathbb{P}(C=K|X =\mathbf{x})
    \end{pmatrix}.
\end{equation}
We now show that the rows have the desired sum.
\begin{equation*}
    \sum_{j=1}^k \mathbf{s}_j(i) = \sum_{j=1}^k \mathop{\mathbb{E}}_{\mathbf{x}\sim \mathcal{D}_i}\mathbb{P}(C=j|X=\mathbf{x})
    = \mathop{\mathbb{E}}_{\mathbf{x}\sim \mathcal{D}_i} \sum_{j=1}^k\mathbb{P}(C=j|X=\mathbf{x})
    = \mathop{\mathbb{E}}_{\mathbf{x}\sim \mathcal{D}_i} 1 
\end{equation*}
Finally, we note that when $\boldsymbol{\pi} = \left(\frac{1}{K},\frac{1}{K},...,\frac{1}{K}\right)$, equation \eqref{eqn:element_expanded} simplifies to
\begin{equation}
        S_{i,j} = \int_{\mathcal{X}}\frac{f_{{i}}(\mathbf{x})f_{{j}}(\mathbf{x})}{\sum_{k=1}^K  f_{{k}}(\mathbf{x})}d\mathbf{x}.
    \end{equation}
In this case $S_{i,j} = S_{j,i}$ and $S$ is symmetric and, therefore, doubly stochastic.$\blacksquare$

\subsubsection{Proof of Lemma \ref{lemma:unique_ortho}}\label{app:general_gramian}

 \begin{duplicatelemma}
    For any two real square matrices $A$ and $B$, $AA^T=BB^T$ if and only if there exists an orthogonal matrix $Q$, such that $A=BQ$.
\end{duplicatelemma}

    We begin with $A=BQ\implies AA^T=BB^T$.
\begin{equation*}
    AA^T = BQ(BQ)^T=BQQ^TB^T=BB^T
\end{equation*}

We now prove $AA^T=BB^T\implies A=BQ$. This proof will take advantage of the \textit{Singular Value Decomposition} (SVD) and the fact that symmetric positive semi-definite matrices have a unique positive semi-definite square root (see \cite{LinAlg_book} Theorem 7.2.6).

We will write the SVDs of $A$ and $B$ as
\begin{align*}
    A &= U_A\Sigma_AV^T_A\\
    B &= U_B\Sigma_BV^T_B,
\end{align*}
where $U_A$, $U_B$, $V_A$, and $V_B$ are orthogonal matrices and $\Sigma_A$ and $\Sigma_B$ are diagonal  matrices with non-negative entries.  Note that 
\begin{equation*}
    AA^T = U_A\Sigma_A\Sigma_AU^T_A=  (U_A\Sigma_AU^T_A)(U_A\Sigma_AU^T_A).
\end{equation*}
We note $AA^T$ is positive semi-definite ($\mathbf{x}^TAA^T\mathbf{x} = || A^T\mathbf{x} ||^2\geq 0$) and $U_A\Sigma_AU^T_A$ is positive semi-definite (because it is similar $\Sigma_A$ which is diagonal with non-negative entries).  This means that $U_A\Sigma_AU^T_A$ is the unique positive semi-definite square root of $AA^T$.  We can similarly prove that $U_B\Sigma_BU^T_B$ is the unique positive semi-definite square root of $BB^T$.  Because $AA^T=BB^T$ and these square roots are unique, we have that 
\begin{equation}\label{eqn:equal_roots}
    U_A\Sigma_AU^T_A = U_B\Sigma_BU^T_B.
\end{equation}

Now consider,
\begin{align}
    A &= U_A\Sigma_AV^T_A\\
    &= U_A\Sigma_AU^T_AU_AV^T_A\label{eqn:multi1}\\
    &= U_B\Sigma_BU^T_BU_AV_A^T\label{eqn:applyroots}\\
    &=U_B\Sigma_BV^T_BV_BU^T_BU_AV_A^T\label{eqn:multi2}\\
    &=BV_BU^T_BU_AV_A^T,
\end{align}
where \eqref{eqn:multi1} and \eqref{eqn:multi2} come from multiplying by the identity ($I = U^T_AU_A=V_B^TV_B $), and  \eqref{eqn:applyroots} comes from applying \eqref{eqn:equal_roots}.  Finally, we note that 
\begin{equation*}
    Q = V_BU^T_BU_AV_A^T,
\end{equation*}
is orthogonal because it is the product of orthogonal matrices.  This gives us
\begin{equation*}
    A = BQ.\blacksquare
\end{equation*}

\subsubsection{Proof of Lemma \ref{lemma:error_bound}}\label{app:posterior_error}

\begin{duplicatelemma}
        The estimate $\hat{\mathbf{y}}(\mathbf{x})$ of the true posterior probability distribution ${\mathbf{y}}(\mathbf{x})$ found using Algorithm \ref{alg:full} when $S$ is diagonally dominant satisfies
    \begin{equation*}
        ||\hat{\mathbf{y}}(\mathbf{x})-{\mathbf{y}}(\mathbf{x})||_\infty \leq 2 \left (\frac{1+\epsilon}{1-\epsilon} \right )\left ( ||\hat{S}-S||_\infty  +\frac{||\hat{\mathbf{q}}(\mathbf{x})-{\mathbf{q}}(\mathbf{x})||_\infty}{||{\mathbf{q}}(\mathbf{x})||_\infty} \right ),\tag{\ref{eqn:posterior_error}}
    \end{equation*}
    where
    \begin{align*}\tag{\ref{eqn:dominance_factor}}
        \epsilon := \max_{i}\frac{\sum_{k\neq i}|S_{i,k}|}{|S_{i,i}|}<1 
    \end{align*}
    is the \textit{dominance factor} which grows closer to zero as $S$ grows more diagonally dominant.
\end{duplicatelemma}

In order to prove this bound, we will use the fact that, for a linear equation $A\mathbf{x} = \mathbf{b}$, the $\hat {x}$ solved for using estimated values $\hat{A}$, $\hat{b}$ satisfies \begin{equation}\label{eqn:condition_bound}
    \frac{||\hat{\mathbf{x}}-\mathbf{x}||_\infty}{||\mathbf{x}||_\infty}\leq\kappa_\infty({A})\left ( \frac{||\hat{A}-A||_\infty}{||A||_\infty} +\frac{||\hat{\mathbf{b}}-\mathbf{b}||_\infty}{||\mathbf{b}||_\infty} \right ).
\end{equation}  See Theorem $2.6.3$ in \cite{condition_number}.  When applied to equation \eqref{eqn:collision}, this bound becomes
\begin{equation}
     \frac{||\hat{\mathbf{y}}(\mathbf{x})-\mathbf{y}(\mathbf{x})||_\infty}{||\mathbf{y}(\mathbf{x})||_\infty} \leq 2\frac{1+\epsilon}{1-\epsilon}\left ( \frac{||\hat{S} - S||_\infty}{||S||_\infty}+\frac{||\hat{\mathbf{q}}(\mathbf{x})-\mathbf{q}(\mathbf{x})||_\infty}{||\mathbf{q}(\mathbf{x})||_\infty} \right ).
\end{equation}
Further, note that $||S||_\infty =1$ because $S$ is row stochastic, and $||\mathbf{y}(\mathbf{x})||_\infty\leq 1$ because $\mathbf{y}(\mathbf{x})$ is a probability vector.  These insights lead to 
\begin{equation}
        ||\hat{\mathbf{y}}(\mathbf{x})-{\mathbf{y}}(\mathbf{x})||_\infty \leq \kappa_{\infty}(S)\left ( ||\hat{S}-S||_\infty  +\frac{||\hat{\mathbf{q}}(\mathbf{x})-{\mathbf{q}}(\mathbf{x})||_\infty}{||{\mathbf{q}}(\mathbf{x})||_\infty} \right ),\label{eqn:kappa_bound_error}
\end{equation}
and in order to prove \eqref{eqn:posterior_error} we need only show that $\kappa_{\infty}(S)\leq 2 \left ( \frac{1+\epsilon}{1-\epsilon}\right )$.

To bound $\kappa_\infty(S)$ (the condition number of the collision matrix $S$ according to the $\infty$-norm),  recall that the condition number is defined as $\kappa_\infty(S)= ||S||_\infty ||S^{-1}||_\infty$. Let $D$ be the diagonal matrix which shares its diagonal elements with $S$, note that $D^{-1}S$ is then the rescaled version of $S$ whose diagonal elements are equal to $1$ and is still diagonally dominant. In that case we may rewrite\begin{equation}
    D^{-1}S = I-E
\end{equation} for some matrix $E$ with zeros on the diagonal. Furthermore, note that $E$ satisfies\begin{equation}
    ||E||_\infty = \max_i  \frac{\sum_{k\neq i}|S_{i,k}|}{|S_{i,i}|} = \epsilon
\end{equation}
by the definition of $\epsilon$ in \eqref{eqn:dominance_factor}.  We may then apply the triangle inequality to find \begin{equation}||D^{-1}S||_\infty = ||I-E||_\infty \leq 1+\epsilon.\label{eqn:straight_norm}\end{equation} 

Now consider that diagonally dominant $S$ is invertible so $D^{-1}S = I-E$ is also invertible.  Because $||E||_\infty=\epsilon<1$ the series $\sum_{i=0}^\infty E^i$ converges and $\left( \sum_{i=0}^\infty E^i \right)\left ( I-E \right ) = I$ (a telescopic series).  Then $\sum_{i=0}^\infty E^i = (I-E)^{-1} = \left ( D^{-1} S \right )^{-1}$.  This allows us to find a bound on the $||\left ( D^{-1} S \right )^{-1}||_\infty$ as follows
\begin{equation}\label{eqn:inverse_norm}
    ||\left ( D^{-1} S \right )^{-1}||_\infty = ||\sum_{i=0}^\infty E^i||_\infty\leq\sum_{i=1}^\infty \epsilon^i = \frac{1}{1-\epsilon},
\end{equation}
where the final equality is a result of $\epsilon < 1$. Combining \eqref{eqn:straight_norm} and \eqref{eqn:inverse_norm} brings us to the condition number bound
\begin{equation}
    \kappa_\infty(D^{-1}S) \leq \left ( \frac{1+\epsilon}{1-\epsilon} \right ).
\end{equation}
Now to get a bound on $\kappa_\infty(S)$ we calculate
\begin{align}\label{eqn:kappa_bound}
    \kappa_\infty(S) = \kappa_\infty(DD^{-1}S) = \kappa_\infty(D)\kappa_\infty(D^{-1}S)\leq 2\left (\frac{1+\epsilon}{1-\epsilon} \right ),
\end{align}
where the final inequality comes from the fact that the $D$ is a diagonal matrix whose elements are bounded between $1/2$ and $1$ (diagonal of a diagonally dominant stochastic matrix).

 Applying  \eqref{eqn:kappa_bound} to \eqref{eqn:kappa_bound_error} results in the desired bound \eqref{eqn:posterior_error}.$\blacksquare$




\subsection{Experiment Details and Model Parameters}\label{app:details_parameters}
All experiments were run using an NVIDIA RTX 4060 GPU. We used the PyTorch python package to create the classifiers for all experiments. Our code may be found online\footnote{\href{https://github.com/JesseFriedbaum/UncertaintyViaCollisions}{https://github.com/JesseFriedbaum/UncertaintyViaCollisions}}.
\subsubsection{Synthetic Dataset Details}\label{app:synth_details}
We elaborate on the dataset generation process for the experiments on synthetic data.  In all Gaussian distributions the covariance matrix was set to the identity matrix $I$ and the means $\mu_k$ where adjusted to change the amount of overlap/uncertainty in the system. We use $\mathbf{1}$ to represent a vector with all elements equal to $1$ where the dimension may be inferred by context.\newline
\textit{{Scenario A}:} The dimension was set at $d=4$, $250$ points were sampled form each class and the number of classes was varied.  We used the following means for each class:
\begin{itemize}
    \item $K=3$ classes: $\mu_1 = 0.25\cdot\mathbf{1}$,   $\mu_2 = -0.25\cdot\mathbf{1}$, and $\mu_3 = 1.25\cdot\mathbf{1}$
    \item $K=4$ classes: $\mu_1 = -0.25\cdot\mathbf{1}$, $\mu_2 = 0.25\cdot\mathbf{1}$, $\mu_3 = 0.75\cdot\mathbf{1}$, and $\mu_4 = 2.5\cdot\mathbf{1}$
    \item $K=5$ classes: $\mu_1 = -0.25\cdot\mathbf{1}$, $\mu_2 = 0.25\cdot\mathbf{1}$, $\mu_3 = 0.75\cdot\mathbf{1}$, $\mu_4 = 2.5\cdot\mathbf{1}$, and $\mu_5 = -1\cdot\mathbf{1}$
\end{itemize}

\noindent\textit{{Scenario B}:}  The dimensionality was fixed at $d=20$ and all experiments used $K=5$ classes and $10,000$ points sampled from each class.  The orientation of the position of the means were selected as $\mu_1 = -3\beta\cdot\mathbf{1}$, $\mu_2 = -\beta\cdot\mathbf{1}$, $\mu_3 = \beta\cdot\mathbf{1}$, $\mu_4 = 5\beta\cdot\mathbf{1}$, and $\mu_5 = 10\beta\cdot\mathbf{1}$.  The value of $\beta$ was adjusted to control the amount of overlap between sets which corresponds to the amount of classification uncertainty.  The most overlap/uncertainty used $\beta = 0.15$, the intermediate overlap/uncertainty used $\beta = 0.25$ and the least overlap/uncertainty used $\beta = 0.35$.\newline
\textit{{Scenario C}:}  This used the same dataset as the most overlap setting from Scenario B.

In these synthetic datasets we have access to the pdfs of the data from each class (the $\mathcal{D}_k$s) and we can use equation \eqref{eqn:element_expanded} to calculate $S$ directly.  We estimate the integral using Monte Carlo integration.  We can also find the true posterior class probability distribution  of any data point $\mathbf{x}$ using
\begin{equation}
    \mathbf{y}_i(\mathbf{x}) = \frac{\boldsymbol{\pi}_if_i(\mathbf{x})}{\sum_{k=1}^K\boldsymbol{\pi}_kf_k(\mathbf{x})},
\end{equation}
where $f_k$ is the pdf corresponding to class $k$. This allows us to implement the Bayes optimal classifier and find the BER, again using Monte Carlo integration.
\subsubsection{Real-World Dataset Details}\label{app:real_details}

We use the MNIST dataset \cite{mnist} as an example of a real-world dataset where the collision matrix and class posterior probability distributions are intuitively understood.  We also use four real-world tabular datasets used in our experiments representing different setting in which there is a clear the possibility of class collisions, indicating the presence of aleatoric uncertainty.

\noindent\textit{German Credit Dataset}: \cite{germancredit_DS}  This commonly used dataset contains information on 1,000 loan applications in Germany labeled by their credit risk.  The opportunity for class collisions in this dataset comes from the fact that many factors not included on a loan application can affect whether the individual pays off the loan.  For example, suppose two individuals submit identical loan applications, but after the loan is granted the first individual loses his job and has a car accident resulting in unforeseen medical expenses.  This could cause the first individual to default on the loan, whereas, the second individual (who did not experience these unfortunate events) pays off the loan, resulting in a class collision. \newline
\noindent\textit{Adult Income Prediction Dataset}:  \cite{adult_DS}  This widely used dataset contains information from the 1994 U.S. census, with individuals labeled by whether their annual income was over \$50,000 ($\sim$\$100,000 in 2024 adjusted for inflation). In this dataset, two people could have identical feature vectors while differing in key ways: for example having majored in very different fields in school or living in totally different parts of the country.  Considering the wide variability between individuals with identical feature vectors we would expect class collisions to be very possible. \newline
\noindent\textit{Law School Success Prediction Dataset}:  \cite{law_DS}  This dataset contains demographic information and  academic records (consisting of undergraduate GPA, LSAT score, and first year law school grades) for over 20,000 law school students labeled by whether or not a student passed the BAR exam.  Here it is very possible that two students with identical features could differ in whether they passed the BAR based off of how much they studied and even sheer luck in guessing answers.  This means that class collisions are possible and aleatoric uncertainty is present. \newline
\noindent\textit{Diabetes Prediction Dataset}: \cite{diabetes_DS}  This dataset contains information on the demographics, health conditions and health habits of 250,000 individuals labeled by whether an individual is diabetic extracted from the Behavioral Risk Factor Surveillance System (BRFSS), a health-related telephone survey that is collected annually by the CDC.  The medical field is notorious for individuals with the same health habits having different medical outcomes indicating that class collisions are possible in this dataset.

\subsubsection{Synthetic Data Model Details and Hyper-Parameters}\label{app:synth_models}

\begin{figure}[t]
    \centering
    \includegraphics[width = \textwidth]{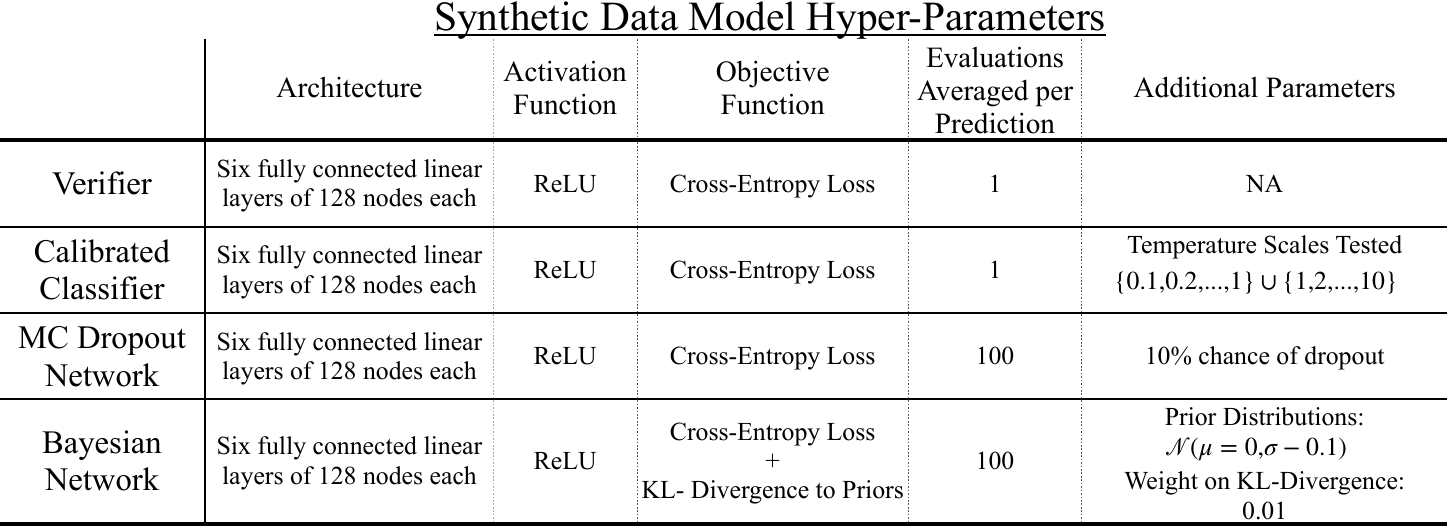}
    \caption{Explanation of the hyper-parameters used to create the verifier $V$ and the various models $M$ used in the baseline approaches on  synthetic datasets.}
    \label{fig:synth_model_table}
    \vspace{-0.25cm}
\end{figure} 

For the experiments on synthetic Gaussian data both classifiers $M$ and pair-wise contrastive models $V$ were implemented using feed-forward  fully connected neural networks.  These models used 6 hidden layers, each containing 128 nodes with ReLU activation functions. We note however, that the number of inputs and outputs does differ between the $M$ and $V$ ($M$ has a $d$ dimensional input and $k$ dimensional output , whereas, $V$ has a $2d$ dimensional input and a $2$ dimensional output).  On synthetic data all models were trained to minimize the cross-entropy loss for a total of 500 epochs, where an epoch involves showing the model $n$ data points. (We do not consider an epoch for the verifier $V$ to involve $n^2$ data points even though there are $n^2$ data pairs in the difference training data.  In this way training the verifier $V$ or a classifier $M$ for one epoch takes approximately the same amount of time.)  Details on the models used are found in Figure \ref{fig:synth_model_table}.  Next we will discuss the computational complexity of the baseline methods for estimating the collision matrix $S$.  The details for estimating  $S$ using equation \eqref{eqn:naive_approx} is described in Algorithm \ref{alg:baseline}.

\begin{algorithm}[t]
\caption{Baseline Collision Matrix Estimation Algorithm}\label{alg:baseline}
\begin{algorithmic}
\STATE {\bfseries Input:} $\mathcal{T} = \{\mathcal{T}_1,\mathcal{T}_2,\cdots,\mathcal{T}_K\}$ partitioned training data and learning rate $\eta$.
\vspace{0.25cm}
\STATE Train classifier $M$ on the training data $\mathcal{T}$ to estimate posterior class probability distributions.
\FOR{$\mathbf{k} \in \{1,2,...,k\}$} 
\STATE Compute $\mathbf{s}_{k} = \frac{1}{|\mathcal{T}_k|}\sum_{\mathbf{x} \in \mathcal{T}_k} C(\mathbf{x})$
\ENDFOR

\STATE return $S$
\end{algorithmic}
\end{algorithm}

\noindent\textbf{Complexity:}  Algorithm \ref{alg:baseline} requires us to train a model $M$ which estimates class posterior probabilities (not just achieve high accuracy) and then evaluate $M$ $n$ times, once for each element of the training data $\mathcal{T}$.  This is dominated by the time required to train the model $M$.  We used three techniques to train a classifier to estimate class posterior probabilities all based on Neural Networks:\newline
\textit{{Calibrated Classifier \cite{calibration_nn}}:}  We use temperature scaling to create a calibrated classifier, motivated by \cite{calibration_nn}, who found this to be the most effective method for calibration.  This method involves splitting the dataset into a training and validation set.  A classifier is first trained to maximize accuracy on the training set.  We then test a variety of temperature scale parameters $T$.  This involves dividing the logits of the model (output immediately before the soft-max layer) by $T$ and then finding the \textit{Expected Calibration Error} (ECE) over the validation set.  The value $T^*$ that leads to the lowest ECE on the validation data is then selected, and in all future evaluations of the classifier logits are divided by $T^*$ before the soft-max layer.  This baseline technique has the least computational expense and only requires the training of a classifier using typical methods and then the calculation of the ECE for a variety of values $T$.  Notably all evaluations of the calibrated classifier are no more computationally complex than evaluations of a regular classifier.\newline
\textit{{Monte Carlo Dropout \cite{mc_dropout}}:}  This method involves training a classifier that includes dropout layers to maximize accuracy on the training data.  During test time we leave the dropout layers functional (contrary to typical usage) which adds randomness to the outputs of the classifier.  During inference we run the input through the network $h$ times where $h$ is a parameter which must be selected.  The final output is then taken as the average of all $h$ outputs.  This technique adds very little complexity to training the algorithm, but does increase the inference costs by a factor of $h$.\newline
\textit{{Bayesian Neural Network (BNN) \cite{BNN_tutorial}}:}  In this method a prior probability distribution is assigned to each parameter of the neural network.  The training data is then used to shift the prior distribution into posterior distributions that better fit the data.  During evaluation we sample all network parameters from the posterior distributions $h$ times.  We then evaluate the input on each of the $h$ networks and average the results to create the final output of the BNN.  This increases inference costs by a factor of $h$ compared to a regular neural network, and as noted in Section \ref{sec:experiments} Figure \ref{img:synthetic-data-results} Scenario 3, requires much longer training times.
\subsubsection{Real-World Model Details Data Model Hyper-Parameters}\label{app:real_models}

\begin{figure}[t]
\begin{center}
\includegraphics[width = 8cm]{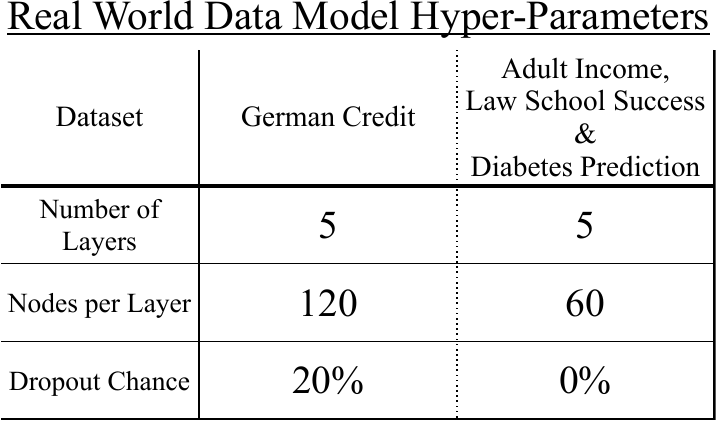}
\end{center}

\caption{Explanation of the hyper-parameters used to create the verifier $V$ on real-world datasets.}
\label{fig:real_hparams}

 For our tests on the MNIST dataset we created the verifier $V$ using a the ResNet 18 architecture \cite{resnet} with two input channels.  To feed two inputs at a time into the classifier, we put each of the images (which only have one channel because they are black an white) into one of the two input channels to the classifier.

\end{figure}

For the real-world tabular datasets, we again used fully connected feed forward neural networks with RELU activation functions.  For these datasets, each network used $5$ hidden layers (each containing $60$ nodes) except for the classifier on the German Credit data which used hidden layers with $120$ nodes and $20\%$ dropout chance on each hidden layer (during training only).  This change was was motivated by the fact that we wished to achieve accuracy on par with common tree based classifiers (random forests and histogram boosted trees).  This was not possible on the German Credit data without augmenting its architecture.  We used an $80-10-10$ train-validate-test data split and stopped training when accuracy on the validation data ceased to increase. See the table in Figure \ref{fig:real_hparams} for a summary of these parameters.

\end{document}